\documentclass[a4paper,11pt]{deepmind}

\usepackage[utf8]{inputenc} 
\usepackage[T1]{fontenc}    
\usepackage{hyperref}       
\usepackage{url}            
\usepackage{booktabs}       
\usepackage{amsfonts}       
\usepackage{nicefrac}       
\usepackage{microtype}      
\usepackage{xcolor}         

\usepackage{graphicx}
\usepackage{subfigure}
\usepackage{amsmath}
\usepackage{amssymb}
\usepackage{algorithmic}
\usepackage{algorithm}
\usepackage[bb=dsserif]{mathalpha}
\DeclareMathOperator*{\argmax}{arg\,max}

\usepackage[natbib=true, style=numeric]{biblatex}
\title{Fast exploration and learning of latent graphs with aliased observations}

\renewcommand{\today}{}

\author[1]{Miguel L\'azaro-Gredilla}
\author[1]{Ishan Deshpande}
\author[1]{Sivaramakrishnan Swaminathan}
\author[1]{Meet Dave}
\author[1]{Dileep George}
\affil[1]{Google DeepMind}
\correspondingauthor{lazarogredilla@google.com}

\begin{abstract}
We consider the problem of recovering a latent graph where the observations at each node are \emph{aliased}, and transitions are stochastic. Observations are gathered by an agent traversing the graph. Aliasing means that multiple nodes emit the same observation, so the agent can not know in which node it is located. The agent needs to uncover the hidden topology as accurately as possible and in as few steps as possible. This is equivalent to efficient recovery of the transition probabilities of a partially observable Markov decision process (POMDP) in which the observation probabilities are known. An algorithm for efficiently exploring (and ultimately recovering) the latent graph is provided. Our approach is exponentially faster than naive exploration in a variety of challenging topologies with aliased observations while remaining competitive with existing baselines in the unaliased regime.
\end{abstract}

\begin{document}

\maketitle

\section{Introduction}
\label{sec:intro}

We consider the problem of efficiently recovering a latent graph from the stream of aliased observations that an agent navigating it perceives. Perceptual aliasing \citep{chrisman1992reinforcement, whitehead1991learning} is a phenomenon that appears when the observations corresponding to different latent states (different graph nodes in our case) are perceived as identical, thus hindering the recovery of the topology of the graph. In addition to dealing with aliasing, to successfully recover the graph's topology, our agent needs to figure out a way to efficiently explore as much as possible of the unknown graph.

\begin{figure}[htbp]
    \includegraphics[width=\linewidth]{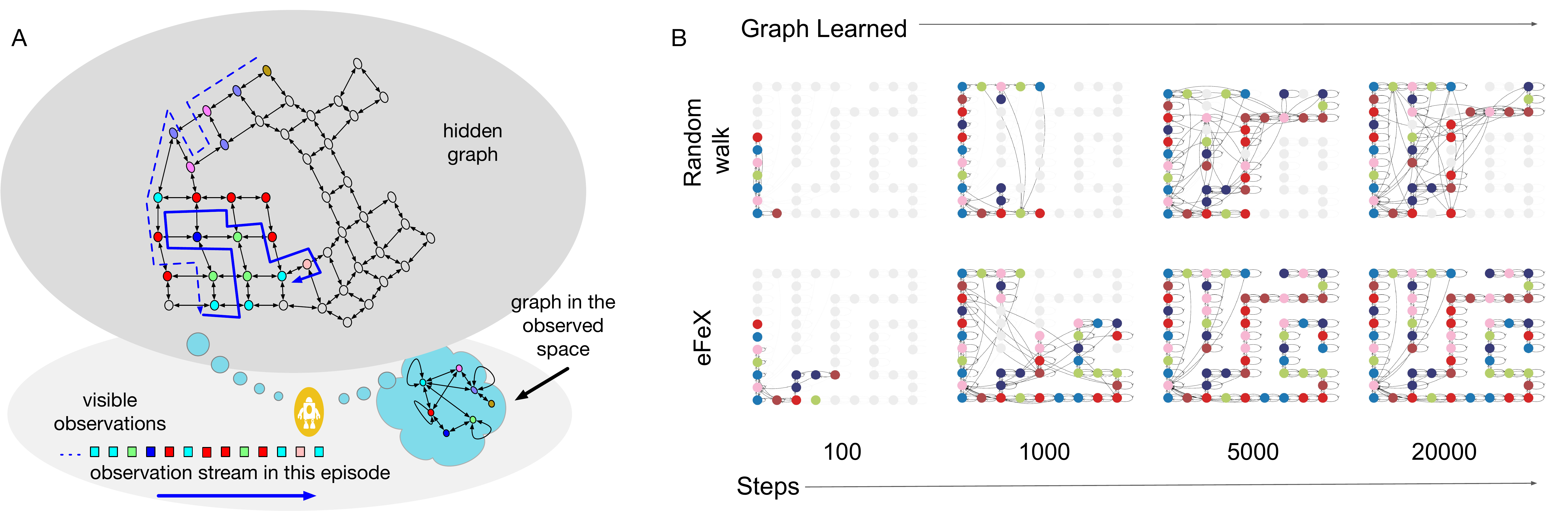}    
    \caption{[Left] An agent explores an environment and receives an \emph{aliased} observation stream. From it, it is able to create a ``mental model'' (dark gray)   that partially matches the true environment, beyond the trivial graph in observed space (light blue). The mental model improves over time. [Right] The agent is informed by its mental model to explore more efficiently (in a different environment).}
    \label{fig:first_page_diag}
\end{figure}

Fig.~\ref{fig:first_page_diag} illustrates this. On the left panel, the agent navigates a latent graph, choosing which actions to take, and receives a stream of categorical observations, encoded as colors. From those observations, it is trivial to build a first-order graph (in light blue background), in which observations correspond to nodes. We tackle the more involved problem of discovering the underlying structure of the latent graph generating those observations (dark gray), where identical observations may correspond to different nodes. On the right panel, on a different latent graph, the effect of using an active exploration policy. Our proposal alternates between inferring a distribution over latent graphs from previous observations, and using that distribution for active exploration using an information gain policy.

The goal of the present paper is to provide an algorithm that recovers an \emph{explicit latent graph} that describes the environment, overcoming the aliasing and exploring actively. We do not concern ourselves with how this graph may be used downstream, but in the following we highlight how multiple tasks are enabled when an agent is able to recover an explicit graph of the environment despite partial observability. We dub our method eFeX (for eFficient eXploration).

First, a graph representation enables efficient planning in the environment. If the origin and target latent states are known, planning (and replanning in the case of stochastic environments) is trivial. Even in the cases in which the agent only has access to the origin and target observations \citep{kaelbling1998planning}, several mechanisms for efficient planning in partially observed environments have been developed \citep{cassandra1994acting, littman1995learning, cassandra1997incremental, spaan2005perseus, meuleau2013solving}. In contrast, standard agents, even if they develop an internal representation of the world, typically will only have a access to a model of the forward dynamics of the world, and not to an explicit graph. Those agents will only be able to plan through forward rollouts and Monte Carlo tree search (MCTS), which is vastly more inefficient than graph planning. Further, the forward models do not need to be consistent and can require extensive training on the type of trajectories that the agent will need to execute at test time to achieve generalization. Reinforcement learning (RL) agents struggle in simple planning tasks, as highlighted in \citep{pasukonis2022evaluating}.

Second, extracting a latent representation corresponds to inducing a complete, explicit map of a partially observed environment from foraging, similar to the task of simultaneous localization and mapping (SLAM), which has been heavily studied in the robotics literature \citep{thrun2002probabilistic, hartley2003multiple, fuentes2015visual}. Neural variants solving this task have also been developed \citep{parisotto2017neural, gupta2017cognitive, chaplot2020learning}. Latent graph extraction is a more general approach, since it does not require observations to be related to the topology, and works for arbitrary topologies, instead of relying on the geometry of 2D or 3D environments. 

Finally, making a full, explicit graph of the environment available to the agent enables advanced capabilities of dynamic modification, transfer learning, and tagging. More explicitly: (a) agents can dynamically modify the graph, marking newly blocked edges as unavailable and being able to replan through a different route; (b) agents can reuse common portions of an extracted graph to learn faster in a new environment, similar to \citep{sharma2021map}; (c) agents can tag latent nodes with specific dynamic information, for instance, where the agent has placed an item that will need later, see also \citep{tong2012gelling}. Because the learned graphs are arbitrary, they are not limited to represent spatial locations, and can conceivably be applied to capture the structure of any sequence.

The contribution is twofold (i) we develop an algorithm that recovers an \emph{explicit graph} of an arbitrary environment (not necessarily Euclidean) from  \emph{aliased} observations, exponentially more efficiently than a random walk; and (ii) we derive an information-theoretic policy for fast exploration similar to \citep{shyam2019model}, but which can be computed in closed form (much faster and without approximation).

\section{Related work}
\label{sec:related}
The potential applications discussed in the previous section connect heavily with the fields of mapping and RL, particularly works that consider partial observability or active exploration. Without being exhaustive, we situate our work in connection with the literature of these fields.

\subsection{Mapping}
Latent graph extraction, when applied in a navigation setting, closely resembles mapping. In a recent line of work \citep{parisotto2017neural, gupta2017cognitive, chaplot2020learning} neural networks (NNs) are used to acquire a map of the environment using an oracle or noiseless ego-motion as supervisory signal, thus providing the exact location to the agent during training. This precludes its application in more realistic settings in which the agent only has access to its local sensing of the environment (which is our setup). SLAM \citep{thrun2002probabilistic, hartley2003multiple, fuentes2015visual, keidar2012robot} considers the more realistic case in which the agent needs to localize itself in addition to performing mapping. However, in SLAM (a) observations are informative about the topology (for instance, the sensors inform of the presence of walls or hallways); and (b) Euclidean geometry is assumed. In contrast, eFeX does not make either of those assumptions. Observations at each location are arbitrary and the topology is defined by a graph, which might have arbitrary jumps between nodes. This allows to model latent states with arbitrary meaning. Other works on mapping \citep{huang2019mapping, jin2021graph} consider states as fully observed, and cannot handle the partial observability of our setup.

\subsection{Markov decision processes (MDPs) and active exploration}
Much of the work in RL is based on the MDP model, in which states are fully observed. Some works \citep{savinov2018semi, zhang2022efficient, lamb2022guaranteed} use latent states, but those can be obtained directly from the observed states via some (possibly learnable) function. Hence, because of the aliasing of our setup, these approaches will conflate all the nodes of the latent graph that emit the same observation.

Work on active exploration has been developed almost exclusively for the MDP model \citep{wilf1989editor, bellemare2016unifying, osband2016deep, shyam2019model, dai2019learning, ritter2020rapid, misra2020kinematic, ecoffet2021first, zhang2022efficient} (see \citep{amin2021survey, yang2021exploration} for a survey), and thus is not directly applicable. The same is true in the case of \emph{options} \citep{machado2017laplacian, machado2017eigenoption}, although these do not tackle the exploration problem directly. In eFeX, we extend \citep{shyam2019model} to the partially observable setting, but introduce some key differences: we provide a closed form expression for the utility, instead of a sampling approximation; this expression is cheap to compute, whereas \citep{shyam2019model} needs to train a full deep NN per sample; we use  value (or policy) iteration to propagate the utility, whereas \citep{shyam2019model} uses Monte Carlo tree search, which can limit the performance when large utilities are distant from the agent; and, fundamentally, we recover an explicit latent graph.

\subsection{Partially observable Markov decisions processes (POMDPs) and aliasing}
POMDPs \citep{sondik1971optimal, kaelbling1998planning} separate the latent space from the observation space through an observation function. Perceptual aliasing can be seen as a particular case of a POMDP in which the observation function is known and collapses all the aliased latent spaces onto the same observation. Obviously, this collapsing is irreversible. Thus, eFeX can be seen as an algorithm to quickly recover the transition function of a POMDP (which can be interpreted as latent graph) in the particular case in which the observation function corresponds to perceptual aliasing \cite{chrisman1992reinforcement, whitehead1991learning}. We ignore the reward and discount factor of the POMDP, since we are only interested in exploration for recovery.

Much of the literature on POMDPs \citep{barto1995learning, mcallester1999approximate, cassandra1994acting, littman1995learning, cassandra1997incremental, pineau2003point, pineau2003point, spaan2005perseus, meuleau2013solving} assumes that the transition function is known (which is rarely the case), skips learning, and tackles the planning problem directly. In contrast, RL approaches do include learning and can be classified in three main categories:

\paragraph{Memory based} A straightforward approach is to augment the observation space by attaching the recent history of observations to it \citep{mccallum1993overcoming, mccallum1996reinforcement, shani2004resolving, shani2005model, bellemare2014skip, bellemare2015count, messias2017dynamic}. Unfortunately, this does not scale whenever long-term memory is necessary to disambiguate the latent state.

\paragraph{Model-based} It is possible to decouple the problem in two parts: a model of the environment is trained to reveal a belief state from a sequence of observations, and that belief is used by an RL algorithm as if it was the observed state \citep{kaelbling1998planning,karkus2017qmdp, ha2018recurrent, igl2018deep, gregor2019shaping, han2019variational, lee2020stochastic, hafner2020mastering}. The belief state is a compact representation of the relevant past, avoiding the need for an ever-growing memory. Particles can be used to capture its uncertainty. Until recently, this was the preferred approach to tackle partial observability.

\paragraph{Model-free} The idea of observing the environment and carrying a state forward can be implemented as a single recursive NN (RNN) \citep{schmidhuber1990making, schmidhuber1990reinforcement, jaderberg2019human, ni2021recurrent}. Training the RNN to both infer the state of the environment and learn to maximize the reward was considered harder than training a model-based system. However, \citep{ni2021recurrent} challenges this view, showing that model-free approaches can perform as well or better than model-based ones.

Even though all of the above approaches contain a latent state (typically in the form a vector of real numbers), none of them produce a graph of the (aliased) environment in which they operate. Although it could be argued that there might be a way to extract a graph of the environment from its latent state, this is yet to be shown. Recent work \citep{pasukonis2022evaluating} shows that simple tasks involving navigation planning in a simple environments with aliasing are not easily solved by deep RL agents.

\section{Method}
\label{sec:method}

\subsection{Background and problem setup}

Consider a categorical state variable $z$ whose values correspond to the nodes of a graph (technically, a multigraph, since it can have parallel edges). At time step $n$, an agent is at node $z_n$, and receives observation $x_n$. The same node always produces the same observation, but multiple nodes can also produce the same observation (i.e., they are \emph{aliased}), so the agent cannot locate itself in the graph from that observation alone. Then, the agent performs an action $a_n$, which (stochastically) determines which directed edge to follow, landing at another node of the graph, $z_{n+1}$. This process repeats, and the agent receives a stream of observation-action pairs ${\cal D}_N \equiv x_1,a_1,\ldots,x_N, a_N$. The goal is to efficiently discover the latent graph, or equivalently, the transition tensor $T$ (with $[T]_{ijk} = P(z_{n+1}=k|z_{n}=j, a_n=i)~~\forall~n$, contains the edges of the graph) and emission matrix $E$ (with $[E]_{ij} = P(x_n=j|z_n=i)~~\forall~n$, contains the emissions associated to each latent node). This corresponds to learning the transition and emission of a POMDP in as few steps as possible, as illustrated in  Fig.~\ref{fig:first_page_diag}.

\begin{figure}[t]
    \subfigure{\includegraphics[width=0.4\linewidth]{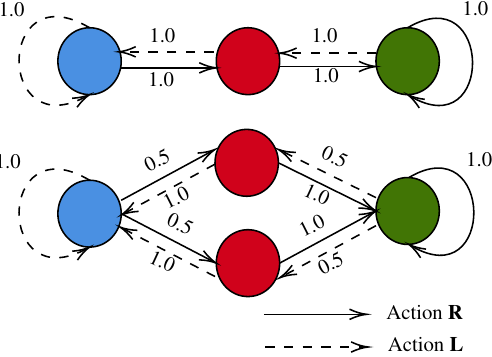}}
    \subfigure{\includegraphics[width=0.59\linewidth]{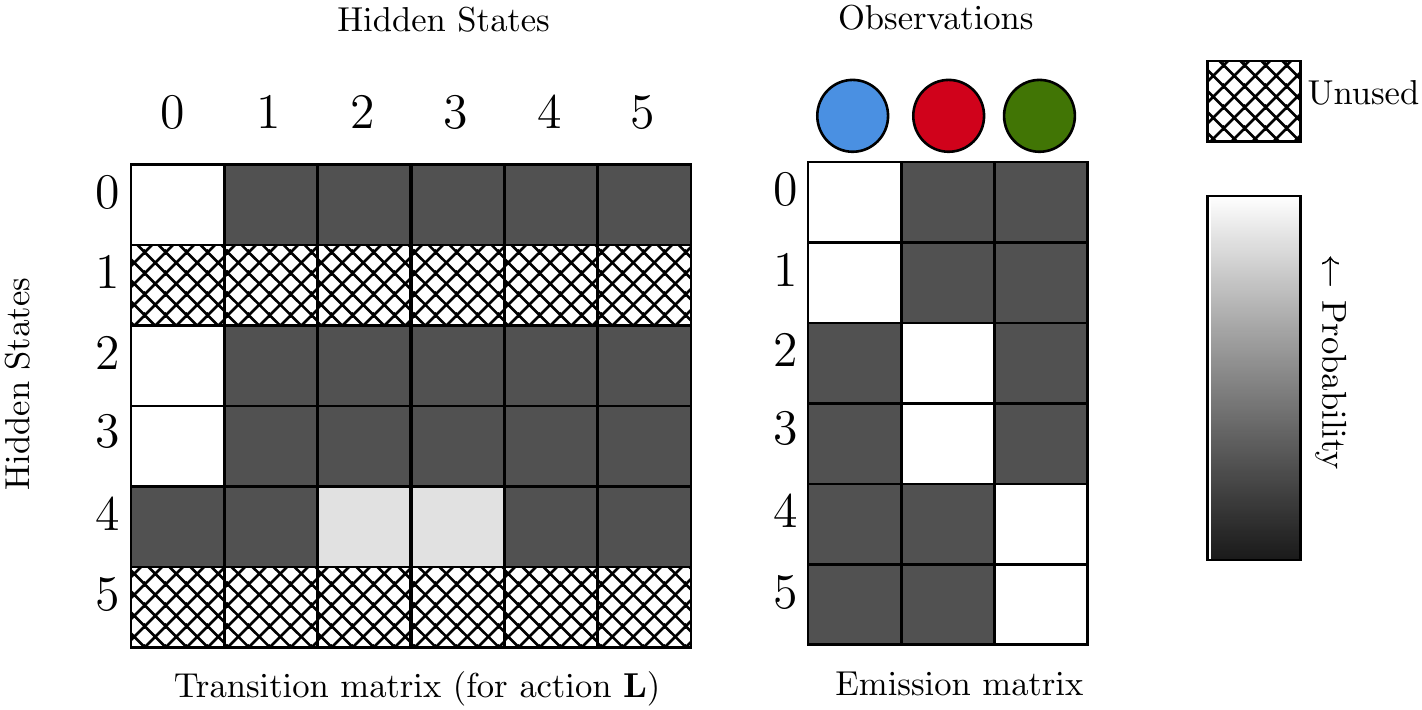}}
    \caption{[Left] Latent graphs with two actions ($L$, $R$). Each edge is labeled with its probability of transition given the state, action. The colors denote the observations. Note that both graphs are degenerate, see Section \ref{sec:aliased_many_topologies} and Appendix \ref{sec:app_degenerate} for more details.
    [Right] Transition matrix (action left) and emission matrix for a CSCG representation of the bottom graph with two clones per observation.}
    \label{fig:degeneracy_and_TE}
\end{figure}

Since the same latent node always emits the same observation, $P(x_n=j|z_n=i)$ is deterministic, all rows of $E$ contain exactly one 1, and its remaining entries are 0. Previous work \cite{george2021clone} suggests to hardcode the emission matrix by allocating a fixed number of latent nodes to each observation (\emph{clones} of that observation), and simply learn the transition matrix via maximum likelihood, letting unused clones be discarded.
See Fig.~\ref{fig:degeneracy_and_TE}[right] for an example of a hardcoded emission matrix where 2 clones are assigned to each observation. Part of the learned tensor $T$ (to be precise, the ``left'' action matrix), is also shown. One of the blue clones and one of the green clones remains unused after learning the transition tensor $T$. This strategy models the sequence as a higher-order network \citep{xu2016representing, lambiotte2019networks}. This model \cite{george2021clone} is called cloned-structure causal graph (CSCG). The likelihood of a graph $T$ is
\begin{equation}
\label{eq:likelihood}
P(x_1,\dots,x_N|a_1,\dots,a_N, T) = 
\sum_{z_1\in {\cal C}(x_1)}\ldots\sum_{z_N\in {\cal C}(x_N)} P_G(z_1)\prod_{n=1}^{N-1}P(z_{n+1}|z_{n}, a_n),
\end{equation}
which can be computed efficiently. The structure in $T$ is discovered via expectation-maximization (EM) \cite{george2021clone}, optionally followed by a few iterations of Viterbi training \cite{brown1993mathematics} to obtain exactly sparse transitions. In fact, running EM to optimize Eq.~\eqref{eq:likelihood} is analogous to running EM to learn an HMM\footnote{While learning HMMs in the presence of perceptual aliasing has been shown to be challenging \citep{chrisman1992reinforcement}, the hardcoding of the emission matrix in \citep{george2021clone} significantly helps learning and accelerates convergence. An explanation is that this model falls within the class of \emph{overcomplete HMMs}, which are known to have favorable properties for learning when the transition matrix is highly sparse and does not contain probable short cycles \citep{sharan2017learning}. Additionally, the sparse structure of $E$ results in significant computational savings during EM \cite{george2021clone}.}. ${\cal C}(x)$ refers to all the latent nodes that emit observation $x$, i.e., all the clones of $x$. 

Our proposal (eFeX) builds on top of CSCGs. CSCGs are a computational model for the hippocampus ---similar in philosophy to the Tolman-Eichenbaum machine \citep{whittington2020tolman}--- that can recover structure from sequences. Structure from multiple domains can be handled, including conceptual, spatial, and others. The present work endows CSCGs with a policy for fast exploration. CSCGs are a probabilistic white-box graphical model, meaning that they recover an explicit, approximate graph of the environment that the agent ``lives'' in. This is particularly convenient because it enables several downstream tasks, as described in Section \ref{sec:intro}.

\subsection{Active exploration based on information gain}

To explore more efficiently, an agent can keep at each step an internal belief (posterior) over $T$, based on the observation-action pairs seen so far. Then it can evaluate, based on its current belief, how informative each potential action would be. Finally, it should take the most informative action.

Let us consider single actions first. Let us say that an agent is at node $z$, it takes action $a$, and lands at node $z'$. How informative is that specific transition? Using the definition of information gain, we have that
$$
\operatorname{IG}(a, z, z') = 
\operatorname{KL}(p(T|a, z, z')||p(T)).
$$ 
The above definition cannot be used directly, since when the agent is at $z$, it needs to decide which action $a$ to take, but it does not know yet where it will land. Instead, we will take its expectation w.r.t.\ $z'$, and additionally marginalize out the transition tensor $T$ according to the current beliefs of the agent. Thus, we  define the utility of action $a$ at node $z$ as
$$
u(z, a) = \int_T \sum_{z'} \operatorname{IG}(a, z, z') P(z'|z,a,T) p(T) dT
 = \operatorname{JSD}\{P(z'|z, a, T)|T\sim p(T)\},
$$
i.e., the Jensen-Shannon divergence among the (infinitely many) distributions $P(z'|z, a, T)$ that can be obtained when sampling from the current belief $p(T)$. This initial definition of utility is identical to the one in \citep{shyam2019model} and has the same motivation. In contrast with that work, we will be able to find a closed-form expression for $u(z, a)$, as opposed to a sampling approximation that requires training a separate neural network (NN) per sample.

\subsubsection{Exact computation of utility for exploration in MDPs (unaliased regime)}
\label{subsec:exactutility}

Let us consider the unaliased case first (which is the only case considered by most of the relevant literature, and in particular, by \citep{shyam2019model}). This corresponds to one clone per observation, or $x_n=z_n$.

For a given action $a$ and node $z$, we can extract the corresponding ``row'' of $T$ into $t_{az}$, a non-negative vector that sums up to one. This vector defines a categorical distribution over the nodes of the graph. We can place a conjugate, symmetric Dirichlet prior with parameter $\alpha$ over this distribution, $t_{az}\sim\operatorname{Dir}(\alpha)$. I.e., $p(t_{az})$ is a distribution over distributions. The posterior after observing one or more transitions is also Dirichlet, so we can keep our belief $p(t_{az})$ conveniently parameterized as a Dirichlet distribution at all times. More precisely, we will keep a ``counts'' vector $c_{az}$ with one entry per destination $z'$. It will store the number of times that we have transitioned from $z$ with action $a$ to each possible destination node. Then, the belief of the agent (posterior) after observing all past transitions, over the transition tensor will be
$p(T|{\cal D}_N) = \prod_{az} p(t_{az}|{\cal D}_N)$,
with
$
p(t_{az}|{\cal D}_N)=
\operatorname{Dir}(c_{az} + \alpha) =
\operatorname{Dir}(b_{az}),
$
where ${\cal D}_N\equiv \left( x_1, a_1,\ldots x_N, a_N \right)$, and $b_{az}\triangleq  c_{az} + \alpha$. Observe that $c_{az}$ can be computed trivially from ${\cal D}_N$, since $z_n = x_n$ and $c_{az}$ are simply counts of observed transitions $z\xrightarrow{a} z'$.
This results in the closed-form utility function
\begin{align}
\label{eq:utility}
\nonumber u(z, a) &= \operatorname{JSD}\{P(z'|z, a, t_{az})|t_{az}\sim \operatorname{Dir}(b_{az})\}
= H({\mathbb E}_{t_{az}\sim \operatorname{Dir}(b_{az})}[t_{az}]) - {\mathbb E}_{t_{az}\sim \operatorname{Dir}(b_{az})}[H(t_{az})]\\
 &=H\Big(\frac{b_{az}}{1^\top b_{az}}\Big)+ \Big(\frac{1^\top (b_{az}\odot\psi(b_{az}+1))}{1^\top b_{az}}\Big)
 -\psi(1^\top b_{az} + 1),
\end{align}
which is one of the main results of this paper. $H(\cdot)$ is the entropy function, $\psi(\cdot)$ is the digamma function, $\odot$ is the elementwise product of two vectors, and $1^\top$ is a row vector of ones of the appropriate length. Details of the derivation are provided in Appendix~\ref{sec:app_proof}. Thus, the exploration utility can be computed directly from the counts $c_{az}$. If we were to only take one action, it should be $\argmax_a u(z, a)$, to maximize the information gain and thus maximally shrink our belief $p(T)$.

\subsection{Reinforcement learning for active exploration}

The above analysis attributes a utility (a one-step information gain) to each node-action pair. When more than one action can be taken, following the one-step maximum information gain may not result in the best total information gain. For instance, a very large utility might appear two steps away from the agent's location, but the local utility of the first step might guide the agent away from it.
Computing the multistep information gain for more than a few steps in infeasible. Instead, we can use the utility as a local reward and obtain a global policy from it by casting the problem as standard reinforcement learning (using some discount factor $\gamma$). This would be exact if the utility did not change. However, as the agent traverses the graph, it captures more information about it, and the utility changes. Therefore, the approximation here is that the utility changes slowly.

We can use value iteration (VI) (or policy iteration) with reward $u(z,a)$ to decide our next action.
Two practical recommendations for computational efficiency are (i) re-run VI after the agent takes a few steps, not just one; and (ii) initialize VI from the result of the previous run. VI requires an action-conditional transition matrix $\overline{T}$, but at any point in time the agent is holding a whole belief $p(T)$ for it, instead of a single point estimate. A simple solution is to set it to its expectation, i.e., use $[\overline{T}]_{az} = \overline{t}_{az} \triangleq {\mathbb E}_{t_{az}\sim \operatorname{Dir}(b_{az})}[t_{az}]=b_{az}/1^\top b_{az},~~\forall~n$.

The utility (if fixed) is propagated optimally, in contrast with \citep{shyam2019model} which requires MCTS. This is in addition to the extra computational cost and approximate utility computation of \citep{shyam2019model}.

\subsection{eFficient eXploration (eFeX) in POMDPs (aliased regime)}
\label{sec:algorithm}

The only input that our active exploration needs in the unaliased regime is the count vectors $c_{az},~\forall~{az}$. These vectors simply record the experienced transitions in the latent space. If we have an estimation of the latent graph describing the environment $T$ (even if approximate), we can use Viterbi decoding to turn the observation sequence $x_1,\ldots,x_N$ into an estimated sequence of hidden states $\hat{z}_1,\ldots,\hat{z}_N$. This allows us to compute the counts  $c_{az},~\forall~{az}$. Putting everything together, we obtain the eFeX algorithm. Fig.~\ref{fig:app_visualization} in Appendix \ref{sec:app_modelvis} shows a visualization of the algorithm, where the utility of each node after a partial exploration is represented by the node size.
\begin{algorithm}[!htb]
    \caption{Efficient exploration (eFeX) for aliased latent graph recovery}\label{alg:efex}
    \textbf{Input} Discount factor $\gamma$, Dirichlet prior $\alpha$, clone allocation $E$ of size $n_E\times n_H$, number of exploration steps $N$.\\
    \textbf{Output} Latent graph in tensor format $T$.
    \begin{algorithmic}[1]
    \STATE{${\cal D}_0\gets(x_1,)$}
    \COMMENT{Init sequence of observations}
    \STATE{$v \gets U[0,1]^{n_H}$}
    \COMMENT{Init all $n_H$ entries randomly between 0 and 1}
    \STATE{$\hat{z}_1 \gets \operatorname{choice}({\cal C}(x_1))$}
    \COMMENT{Init $z_1$ to a random clone of $x_1$}
    \STATE{$c_{az} \gets 0,~\forall_{az}$}
    \COMMENT{Init $c_{az}$ to a vector of $n_H$ zeros}
    \FOR{$n$ in $1,\ldots,N$}
        \STATE{$b_{az} \gets c_{az} + \alpha,~\forall_{az}$}
        \STATE{$u(z, a) = \operatorname{utility}(b_{az}),~\forall_{az}$}
        \COMMENT{Compute utility using \eqref{eq:utility}}
        \STATE{$\overline{t}_{az} \gets b_{az}/1^\top b_{az},~\forall_{az}$}
        \COMMENT{Compute mean transition tensor}
        \REPEAT
            \FOR{$z$ in $1,\ldots,n_H$}
                \STATE{$[v]_z \gets \max_{a} \overline{t}_{az}^\top((1-\gamma)u(z, a)+\gamma v)$}
                \COMMENT{Run value-iteration}
            \ENDFOR
        \UNTIL{convergence of $v$}
        \STATE{$[\pi]_z = \argmax_{a} \overline{t}_{az}^\top((1-\gamma)u(z, a)+\gamma v)$}
        \COMMENT{Recover optimal policy $\pi$}
        \STATE{$a_n \gets [\pi]_{\hat z_n}$}
        \STATE{$x_{n+1} \gets \operatorname{Execute}(a_n)$}
        \COMMENT{Take approximate best action, receive observation}
        \STATE{${\cal D}_n\gets{\cal D}_{n-1} \cup (a_n, x_{n+1})$}
        \COMMENT{Grow sequence of observations}
        \STATE{Use EM + Viterbi training to obtain the transition tensor $T$ from ${\cal D}_n$ (with pseudocount $\alpha$)}
        \STATE{Use Viterbi with $T$ and $E$ on ${\cal D}_n$  to obtain a decoding $\hat{z}_1,\ldots,\hat{z}_{n+1}$}
        \STATE{Use $\hat{z}_1,a_1\ldots, a_n, \hat{z}_{n+1}$ to compute $c_{az},~\forall_{az}$}
    \ENDFOR
    \end{algorithmic}
\end{algorithm}

\section{Experiments}
\label{sec:experiments}

We evaluate Algorithm \ref{alg:efex} (eFeX) on different topologies and under different levels of aliasing. We set the Dirichlet prior $\alpha=2\times 10^{-3}$, and the discount parameter $\gamma=0.9999$. The number of clones per observation just needs to be ``large enough'', and we set it to twice its average ground truth value. Using a larger than necessary number helps learning in latent variable models, see \citep{buhai2020empirical}, and \citep{xu2018benefits} for the specific case of EM learning. Results are not very sensitive to parameter choices.

\subsection{Fully observed graphs (unaliased regime)}
\label{sec:nonaliased}

\begin{figure}[!htbp]
\centering
\includegraphics[width=0.8\linewidth]{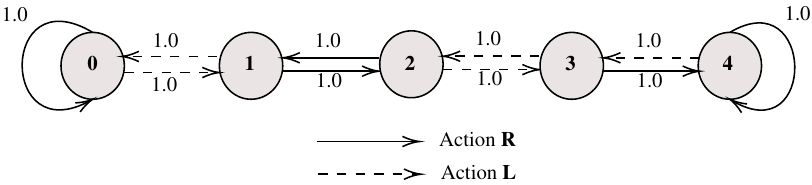}\\
\includegraphics[width=0.45\linewidth]{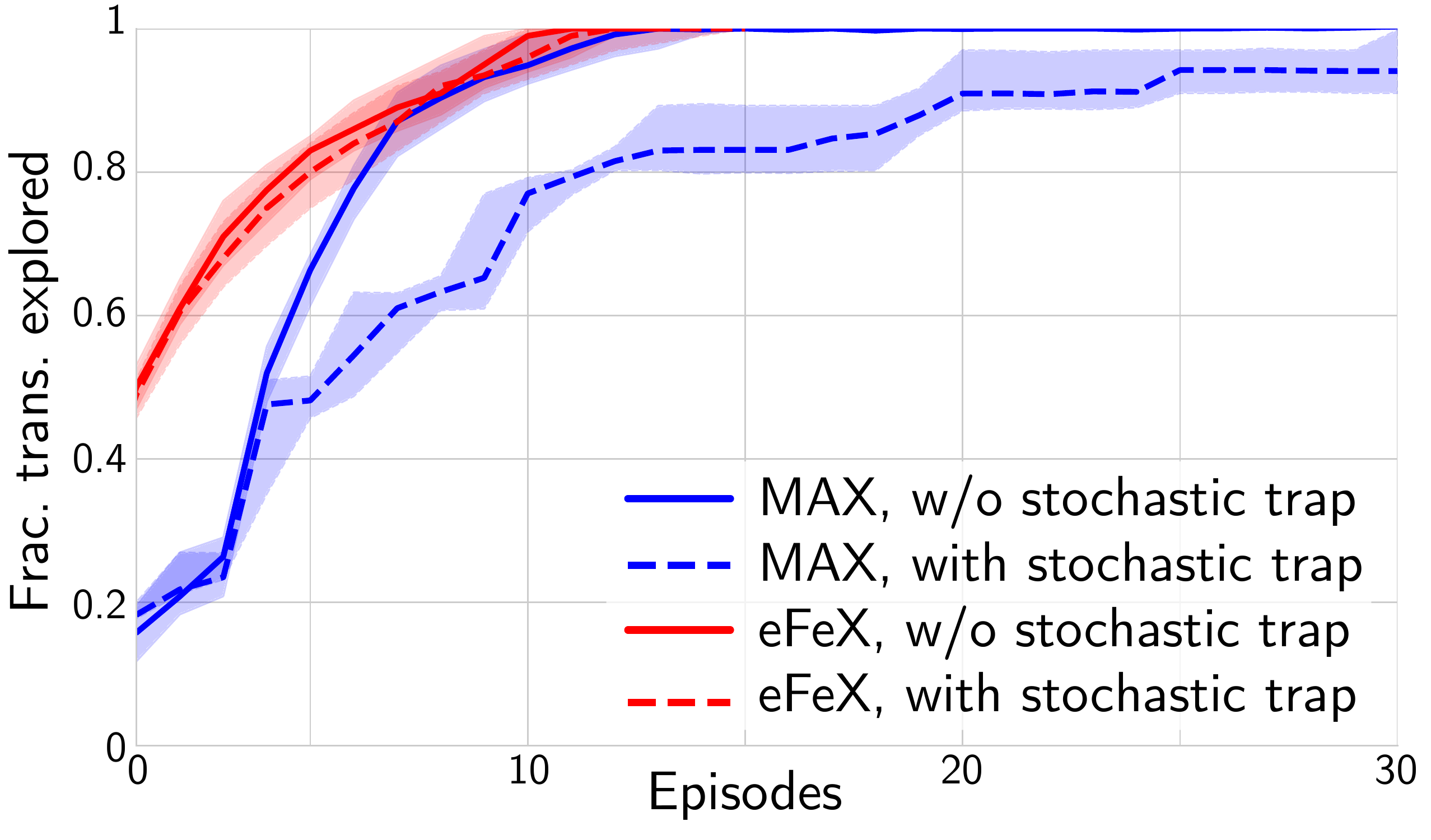}
\includegraphics[width=0.5\linewidth]{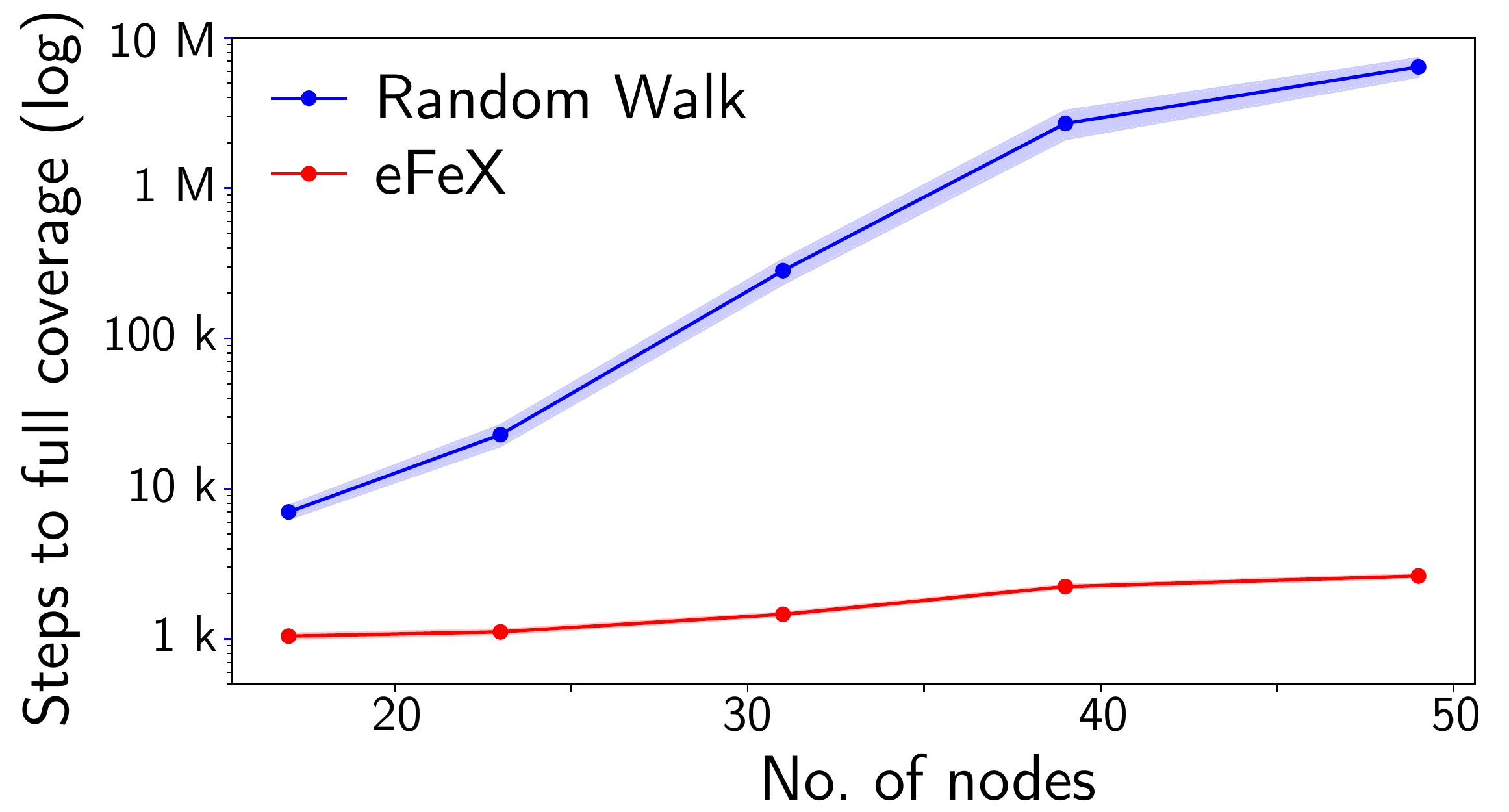}
\caption{[Top] Chain for $L=5$. [Left] Coverage for $L=50$. The stochastic trap barely affects the performance of eFeX. [Right] Scaling with the size of the ``maze'' topology, under the random and eFeX policies. eFeX is exponentially more efficient.  Averaged over 100 runs. 95\% confidence intervals provided.}
\label{fig:chain1d_scaling}
\end{figure}

We consider first the case in which the environment has no aliasing. This makes it easier to compare with existing work in the literature of active exploration. In particular, we consider here the chain graph environment (see Fig.~\ref{fig:chain1d_scaling}[top]) from \citep{shyam2019model}. It was posed originally in \citep{osband2016deep} as a simplification of the ``River Swim'' problem from \citep{strehl2008analysis}. The environment consists of a chain of $L$ nodes, each emitting a unique observation. Each node has two outgoing edges connecting it to its neighbors. What makes it tricky is that the labels of those edges (the actions that traverse the chain, noted as {\tt L} and {\tt R} in Fig.~\ref{fig:chain1d_scaling}[top])) are randomly flipped along the chain, so both a random policy and a policy that always repeats the same action will get trapped in a loop and not traverse the full chain. The end nodes of the chain are simply a self-loop. The problem is posed as a sequence of episodes of length $L+9$, with the agent starting each episode at the node labeled as ``1''. This means that  to fully explore the chain, the agent must choose the adequate action at almost every step.

A more difficult variant introduced in \citep{shyam2019model} incorporates a ``stochastic trap'' at node ``0'': both the {\tt L} and {\tt R} actions will remain in node ``0'' or transition to state ``1'' with probability 0.5 (not pictured). The stochastic trap will trick models that conflate \emph{risk} (unlearnable unknowns) with \emph{uncertainty} (learnable unknowns). Such models will be lured by the randomness of results experienced at node ``0'', thinking it is uncertainty, and that they should keep exploring it to reduce it, when it is actually unavoidable risk. Thus, such models will make little progress in exploring the rest of the chain. 

We compare eFeX with the model-based active exploration from \citep{shyam2019model}, called MAX, which in turn beats deep Q-network (DQN) approaches such as exploration bonus DQN \citep{bellemare2016unifying} and bootstrapped DQN \citep{osband2016deep}. As shown in Fig.~\ref{fig:chain1d_scaling}[left], without stochastic trap, MAX only needs 18 episodes (including 3 episodes of warm-up time) for exploring all the transitions in a 50-state chain (much fewer than deep Q-learning, not shown here). But eFeX is faster, only needing 11 in the median case. The gap becomes much larger once the stochastic trap is activated, with eFeX barely slowing down. This also shows that eFeX, like MAX, is not conflating risk and uncertainty. The superiority of eFeX over both deep Q-learning and MAX is not surprising given that MAX beats deep Q-learning and eFeX improves over MAX in terms of utility computation and propagation (see Sections \ref{sec:related} and \ref{sec:method}). 

\subsection{Partially observed graphs, multiple topologies (aliased regime)}
\label{sec:aliased_many_topologies}

In the previous section we measured only the exploration capability (fraction of transitions explored), since without aliasing, graph recovery is trivial for a fully explored graph. With aliasing, the quality of the recovery can be poor even after full exploration. Furthermore, in a stochastic graph, some transitions may have a vanishingly small probability, so missing them should not affect the measure much. We use three measures for latent graph recovery: \emph{expected log-likelihood} (measures the quality of the recovered graph, and is therefore affected by adequate exploration, but is not very intuitive), \emph{weighted coverage} (downweights the importance of less likely edges, but measures only exploration, ignoring the quality of the recovered graph), and \emph{precision} (measures the quality of the recovered graph, is affected by the quality of the exploration, and is easy to interpret). We report the latter here, which measures the precision of a mapping from the decoded vertices of the recovered graph to the ground truth vertices, which must be between 0 and 1. Mathematically, given the ground truth latent sequence $\{z_n\}_{n=1}^N$ and the one recovered by eFeX, $\{\hat{z}_n\}_{n=1}^N$ under a walk in which actions are chosen uniformly at random, the precision is
$$
    \operatorname{PREC} = \lim_{N\to\infty}\mathbb{E}_{a_1,\ldots,a_N\sim \text{U}}\Big\{\frac{1}{N} 
    \sum_i \max_j \sum_n \mathbb{1}[\hat{z}_n=i]\mathbb{1}[z_n=j]\Big\}.
$$
where $\mathbb{1}[\cdot]$ evaluates to 1 if the inner expression is true and 0 otherwise. This measure captures both the quality of the exploration and the graph recovery.
We include results with all measures in Appendix \ref{sec:app_exp}. For a detailed explanation of each performance measure, see Appendix \ref{sec:app_performance}.

Two different latent graphs may produce the same action-conditional distribution over their emissions. We call such latent graphs \emph{degenerate}. An example of two degenerate graphs is shown in Fig.~\ref{fig:degeneracy_and_TE}[left]. One graph was obtained from the other by splitting a clone into two. The types of degeneracy are described in Appendix \ref{sec:app_degenerate}. Since degenerate graphs cannot be disambiguated from a sequence of observations, they are all considered equally valid solutions to the problem of latent graph recovery. In fact, the measures described above are designed to be insensitive to degeneracy. Most downstream applications will not be affected by degeneracy. If necessary, it is possible to bias the solution towards one particular type of degeneracy, e.g., by merging equivalent clones.

\begin{figure*}[t]
\centering
    \includegraphics[width=0.9\linewidth]{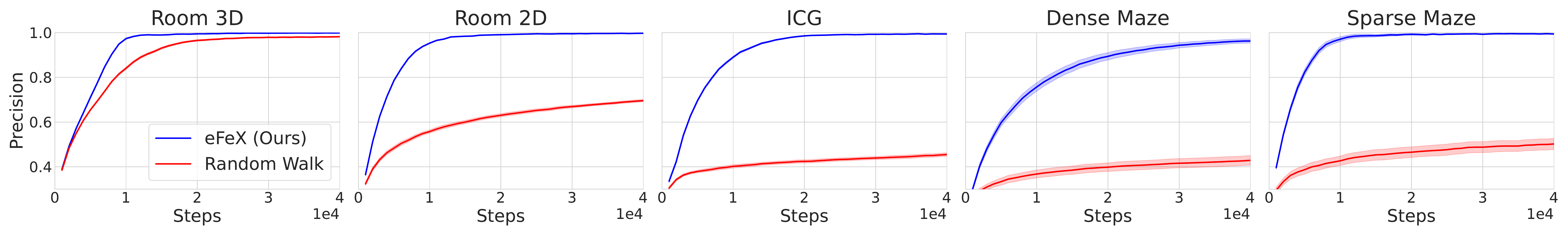}\\
    \hspace*{0.2cm}
    \includegraphics[height=2.6cm]{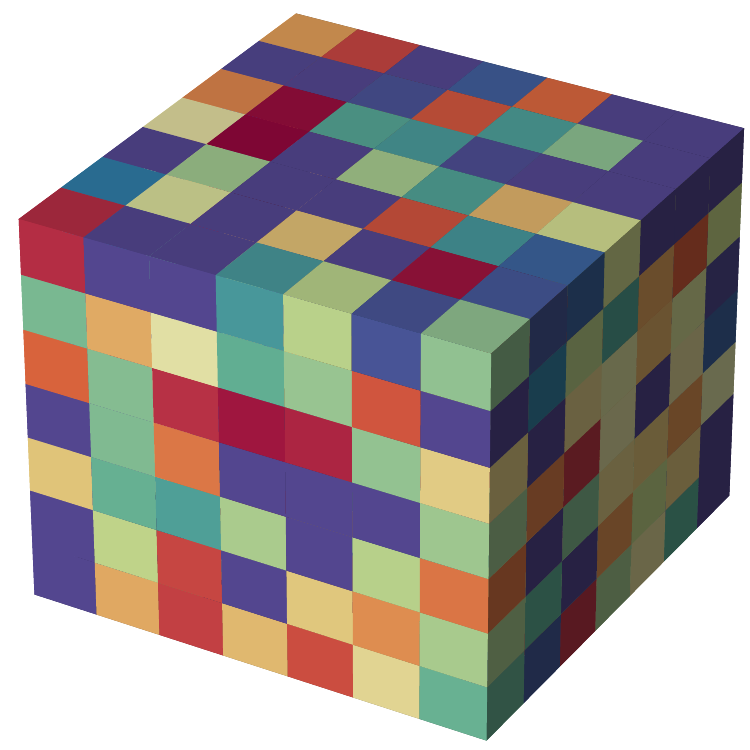}\hspace{0.34cm}
    \includegraphics[height=2.6cm]{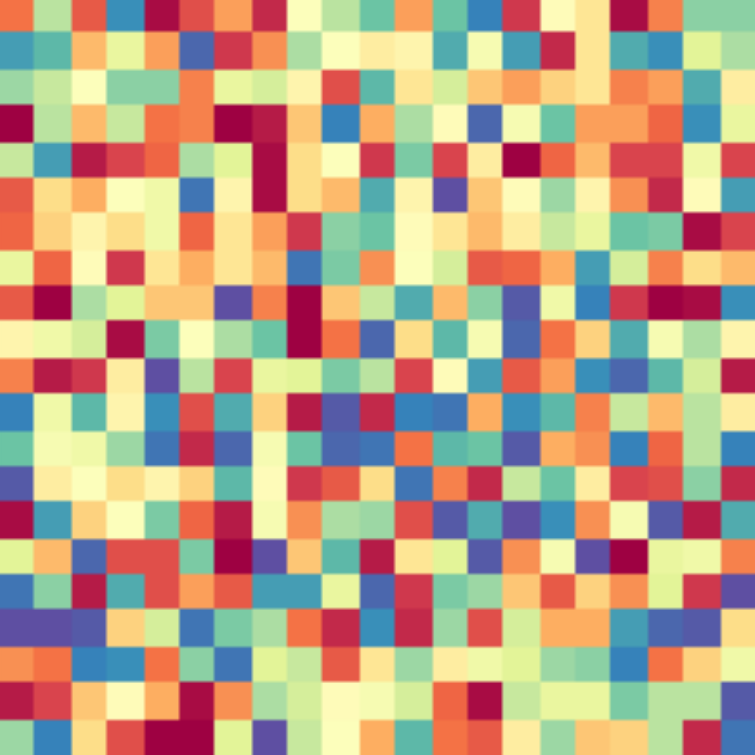}\hspace{0.2cm}
    \includegraphics[height=2.6cm]{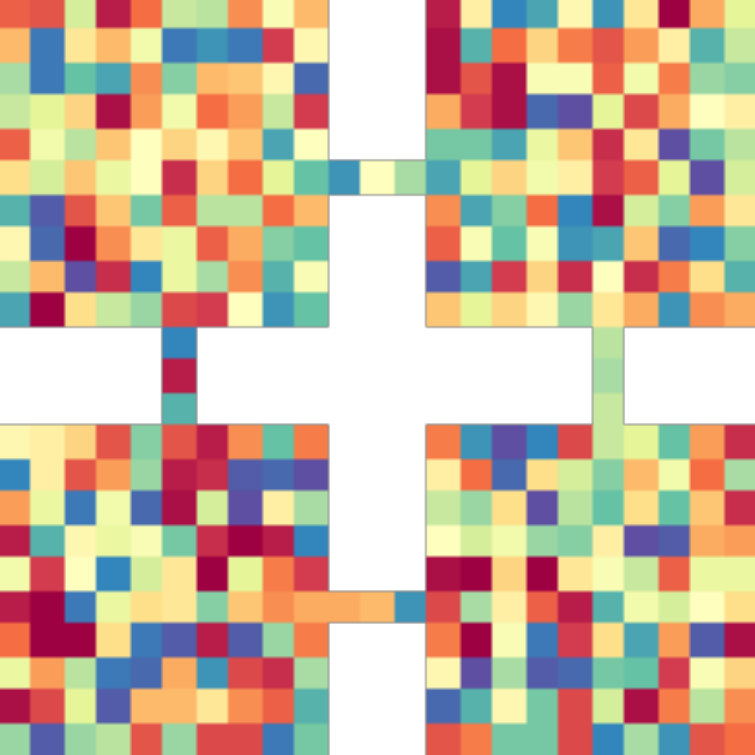}\hspace{0.2cm}
    \includegraphics[height=2.6cm]{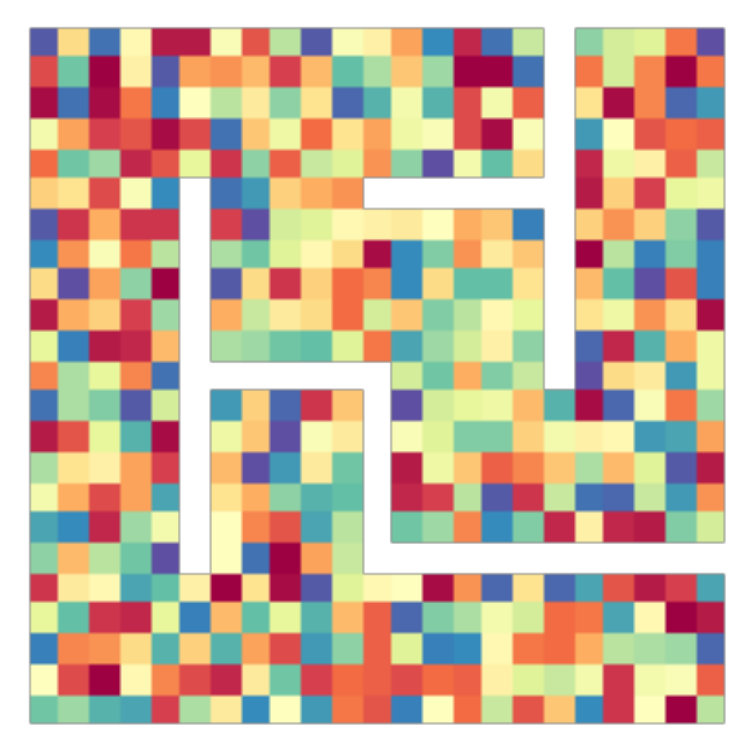}\hspace{0.15cm}
    \includegraphics[height=2.6cm]{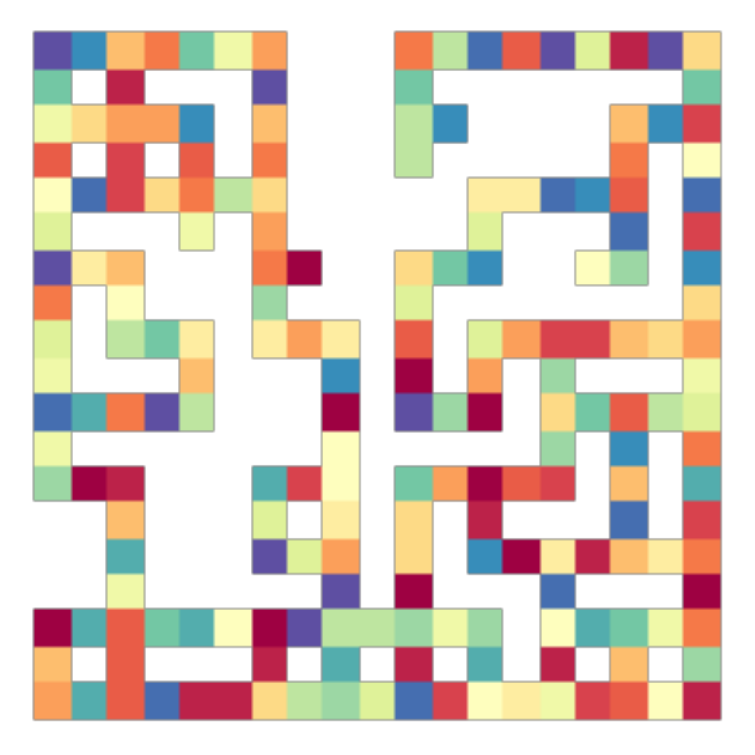}\hfill
    \caption{[Top] Precision for each topology, as a function of the number of steps of the agent in the environment. Higher is better. Results are averaged over 40 runs. Bands indicate 95\% confidence intervals for the averages. Random policy shown in blue, eFeX in red. Environment parameters $\operatorname{UF}=0.1$ and episode length 100. [Bottom] Exploration environments. Colored locations are accessible, and white regions are inaccessible. The colors represent the (categorical) observation that the agent receives at each location.}
    \label{fig:topologies}
\end{figure*}

The topologies that we will be testing are shown in Fig.~\ref{fig:topologies}[bottom], with colors representing the (repeated) observations. In detail (from left to right in the figure):

\paragraph{Regular grids} A $7\times 7\times 7$ 3D grid and a $21\times 21$ 2D grid. With respectively 6 and 4 actions, allowing the agent to traverse to the adjacent cells (and further due to slippage). Boundaries contain self-loops, so the agent will remain in place when attempting to exit the topology. These structures do not contain obstacles or bottlenecks that hinder exploration, so a random policy will explore them fairly quickly, although it might struggle to capture the last few nodes.

\paragraph{Interconnected grids (ICG)} Four 2D grids of size $10\times 10$ connected by corridors of length 3. Each grid forms a cluster of highly connected nodes, whereas connecting across grids is more challenging. These are similar to ``lollipop'' graphs discussed in \citep{chandra1989electrical} which, as noted in the reference, mimic real world connectivity patterns. Such graphs require the agent to deliberately go through the corridors in order to get to different parts of the graph. Boundaries are self-looped as above.

\paragraph{Dense maze} Dense mazes are generated by running a depth-first search on a 2D grid, and then up-sampling it by a factor of five to make the corridors five times as thick as the walls, with the final result being a $21 \times 21$ grid with gaps. Boundaries are self-looped. These mazes demand that the agent performs long sequences of suitably correlated actions to fully recover the environment, which in turn necessitate a good model of how the environment explored so far reacts to actions. Boundaries are self-looped as above.

\paragraph{Sparse maze} Sparse mazes are generated on a background of size $21 \times 21$ using the open source ``LabMaze'' library introduced in \cite{beattie2016deepmind}. Boundaries are self-looped as above.

The degree of aliasing of an environment is measured by the \emph{unique fraction}, $\operatorname{UF} = \frac{\text{\# unique observations}}{\text{\# nodes in graph}}$. No aliasing corresponds to $\operatorname{UF}=1$, and recovery is harder as this number shrinks. The agent has access to episodes in the environment of length 100, where it can act. At the beginning of each episode the agent is reset to a ``home'' location in a corner of the topology. Despite this, Fig.~\ref{fig:topologies}[top] shows that eFeX is able to achieve almost perfect precision in an environment with significant aliasing $\operatorname{UF}=0.1$. This implies that both recovery exploration are almost perfect. In contrast, the random walk explores the environment much more slowly, almost grinding to a halt in the more complicated topologies. In Appendix \ref{sec:app_exp} we include more results with other degrees of aliasing and all the mentioned performance measures, and introducing stochasticity via \emph{slippage}. Slippage makes the environment repeat a basic action (such as {\tt left}), $k$ times, with $P(k) = (1-P_\text{slip})P_\text{slip}^{k-1},~~k\geq 1$ when the agent executes it once. The agent has no access to the amount of slippage that has occurred, just to the observation, as always, which due to aliasing will not resolve the amount of slippage either. This makes the environment even more challenging to learn.

Finally, we consider an even more challenging topology. We take the ``sparse maze'' depicted in Fig.~\ref{fig:topologies}[bottom-right], and set $\operatorname{UF}=0.1$. Actions that would exit the topology result in the agent staying in place 90\% of the time, and teleporting ``home'' 10\% of the time. A random walk will require \emph{exponential time} to cover the graph, whereas eFeX is closer to linear in efficiency, see Fig.~\ref{fig:chain1d_scaling}[right].

Most of the methods in the exploration literature cannot be applied in this case due to aliasing (they would conflate all the locations with the same state), nor would they recover the latent graph.

\section{Discussion}
\label{sec:discussion}

We have introduced eFeX, an algorithm for fast recovery of aliased latent graphs. In the simpler, fully observed setting, we can establish direct comparisons with recent, state-of-the-art algorithms, such as the model-based MAX \citep{shyam2019model}, the deep-Q-learning-based exploration bonus DQN \citep{bellemare2016unifying}, and bootstrapped DQN \citep{osband2016deep}. eFeX is competitive with these baselines, but graph extraction is trivial.

In the aliased regime, eFeX is, to the best of our knowledge, the first algorithm able to recover the latent graph through the use of an active policy aimed at minimizing the number of steps in the environment. We show empirically that when the topology becomes challenging, eFeX can be exponentially better than a random policy. As shown in Appendix \ref{sec:app_exp}, both aliasing and stochasticity reduce the speed at which we can recover a latent graph, since the problem is harder. But eFeX does not break as we introduce aliasing and stochasticity, instead degrading gracefully.

Although we do not delve into it in this work, latent graph recovery in the presence of partial observability (precisely, aliasing) has multiple downstream applications, particularly in the field of RL. The resulting model is a directed multigraph representing the environment, which is easily inspectable and modifiable. Having an explicit graph of the environment enables or simplifies multiple tasks: efficient planning (without rollouts or MCTS), dynamic adaptation of the graph, human-interpretable map extraction, location tagging, shortcut finding, transfer learning of graph portions for fast explorations of related novel environments, etc.

\medskip
\printbibliography

\newpage
\appendix
\onecolumn
\section{Algorithm visualization}
\label{sec:app_modelvis}

\begin{figure}[htbp]
    \centering
    \subfigure{\includegraphics[width=0.25\linewidth,trim={85mm 40mm 85mm 30mm},clip]{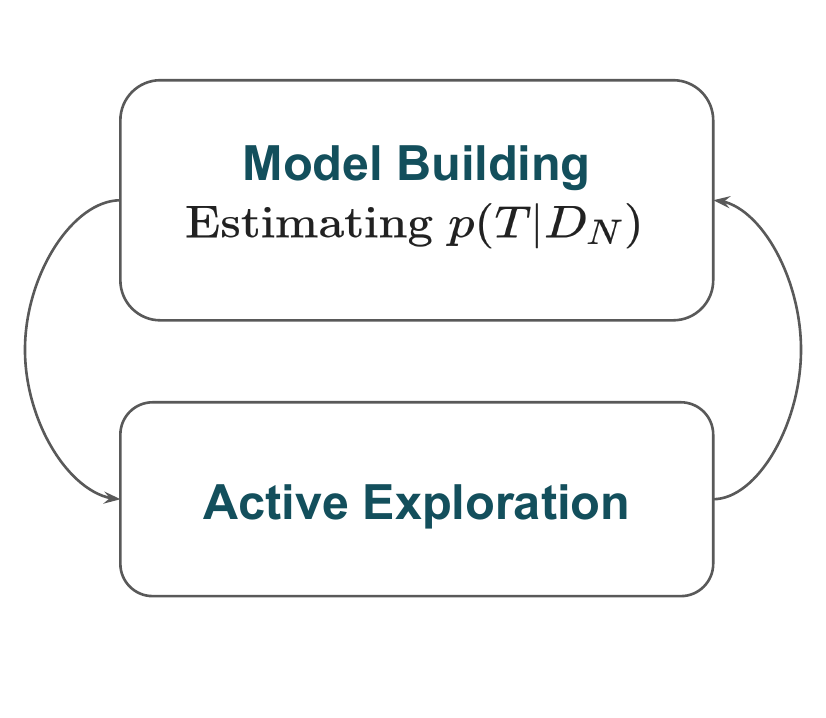}}
    \hspace{0.7cm}
    \subfigure{\includegraphics[width=0.25\linewidth]{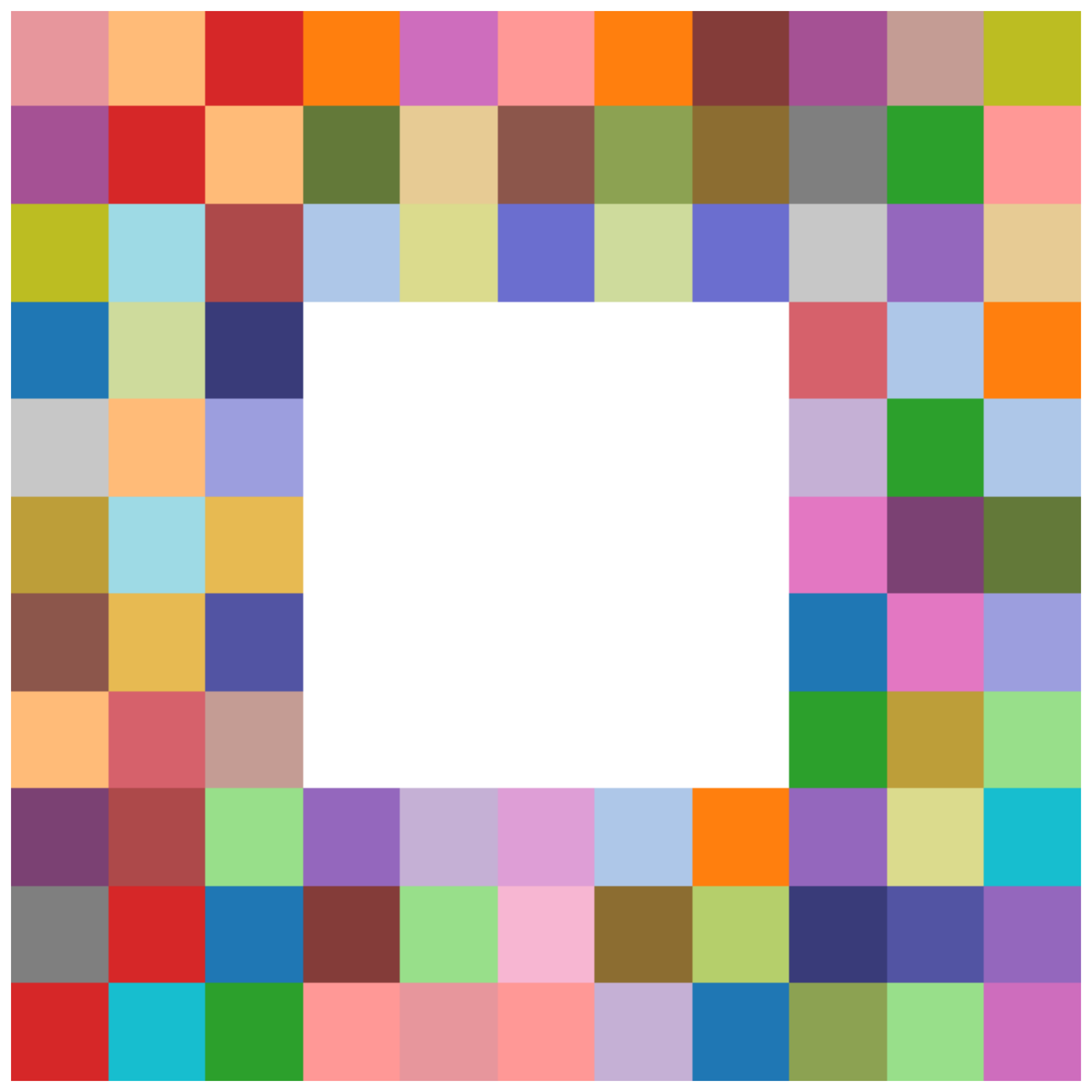}}
    \hspace{0.7cm}
    \subfigure{\includegraphics[width=0.21\linewidth]{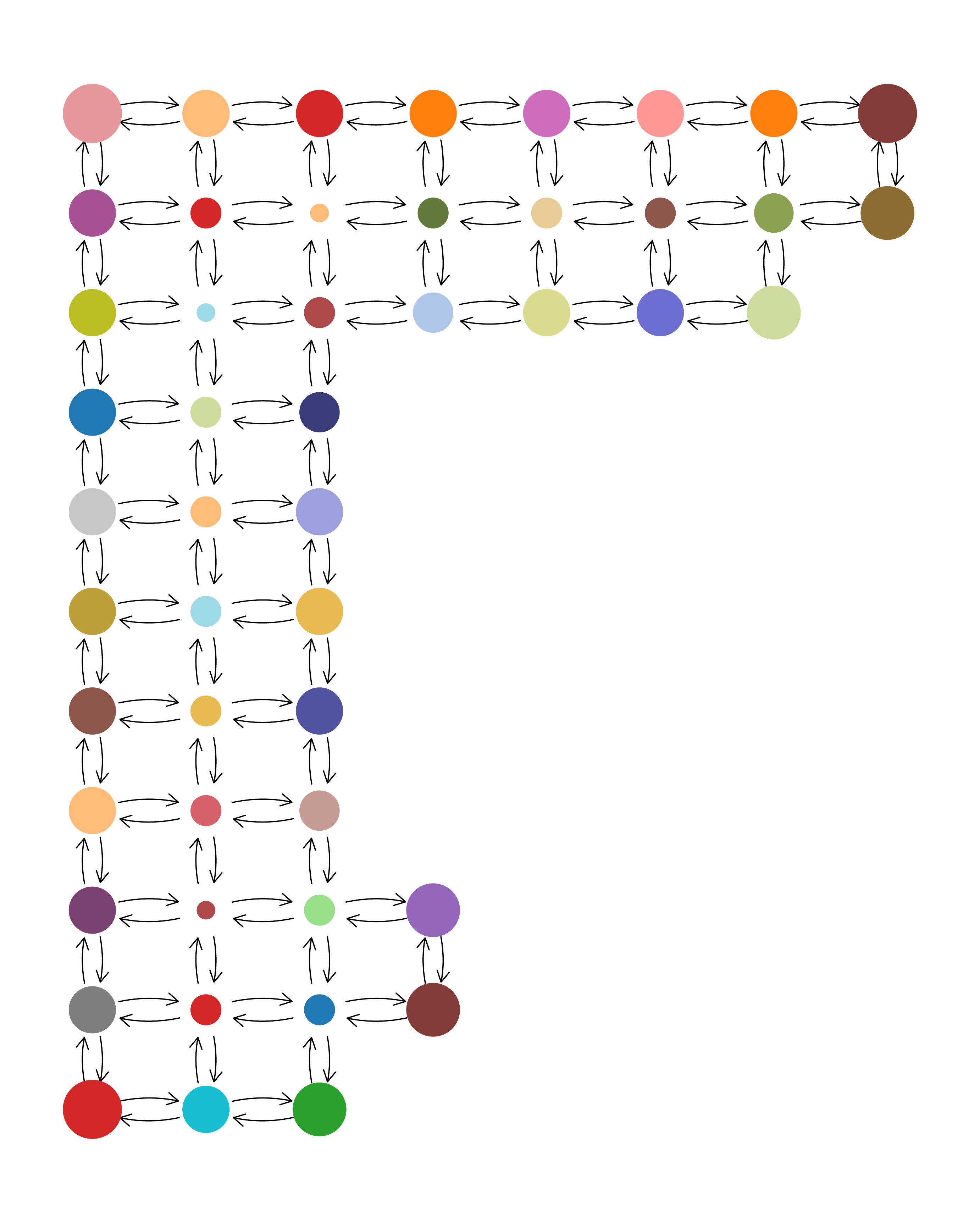}}
    \caption{[Left] The eFeX algorithm alternates between estimating a distribution over stochastic graphs given the observed data $p(T|{\cal D}_N)$ and using that distribution to decide which action to take next (active exploration) 
    [Middle] Ground truth environment 
    [Right] Learned partial graph, with node size scaled according to the utility, $\max_a u(z, a)$ for each node $z$.}
    \label{fig:app_visualization}
\end{figure}

\section{Performance measures}
\label{sec:app_performance}

To properly specify the latent graph recovery problem, we need to provide a quantitative measure of the quality of an agent's solution. We will refer to the ground-truth graph as $T^\text{GT}$ and to the agent's recovered graph as $T$.

\paragraph*{Expected log-likelihood}
The expected log-likelihood of $T$ under $T^\text{GT}$ for given observations $\mathbf{x}\equiv x_1,\ldots,x_N$ and $\mathbf{a}\equiv a_1,\ldots,a_N $ is
$$
{\cal L}_{T^\text{GT}}(T) = \lim_{N\to\infty}\mathbb{E}_{\mathbf{a}\sim \text{U}}\Big[\sum_{x_1,\ldots,x_N}
P(\mathbf{x}|\mathbf{a}, T^\text{GT})
\log P(\mathbf{x}|\mathbf{a}, T) \Big],
$$
where $\text{U}$ generates sequences by sampling from the uniform distribution over the available actions at each time step, i.e., uses a random policy. This measure is always negative, and higher is better. A value of ${\cal L}_{T^\text{GT}}(T)=0$ can only be achieved when $T$ is perfectly recovered and actions have deterministic results. In general, ${\cal L}_{T^\text{GT}}(T)\leq{\cal L}_{T^\text{GT}}(T^\text{GT})$, and reaching the upper bound implies perfect recovery.

Observe that this definition has a very convenient property: an agent traversing the environment taking a random action at each time step can get an arbitrarily good approximation to ${\cal L}_{T^\text{GT}}(T)$ by traversing the graph for long enough and with enough restarts. I.e., access to the true $T^\text{GT}$, or to the true latent state of the graph during traversal, is not necessary to compute a good approximation.

This is both an information-theoretic measure that quantifies the quality of the recovery, and the quantity that we optimize w.r.t.~$T$ during eFeX training to estimate the model. The disadvantage is that the value does not gives an absolute sense of how good the recovery is, or how well the agent knows its situation in the graph.

\paragraph*{Weighted coverage} To assess whether an active policy $\Pi$ (such as the dynamic policy in eFeX) is efficiently exploring a latent graph $T^\text{GT}$, we may want to measure how fast new edges of $T^\text{GT}$ are being traversed as we take more exploration steps. We define the weighted coverage as follows
\begin{equation}
    \operatorname{WCOV}_{T^\text{GT}}(\Pi, N) = \frac{\sum_{n=1}^N P_{T^\text{GT}}(z_{n+1}|z_n, a_n)
    \mathbb{1}[(z_n, a_n, z_{n+1})\neq (z_k, a_k, z_{k+1}) ~~\forall_{k<n}]}
    {\sum_{i,j,k} P_{T^\text{GT}}(z_{n+1}=k|z_{n}=j, a_n=i)},
    \label{eq:wcoverage}
\end{equation}
where the indicator function $\mathbb{1}[\cdot]$ evaluates to 1 if the expression inside the brackets is true and to 0 otherwise.
The numerator iterates over the ground truth triplets $(z_n, a_n, z_{n+1})$ that define the edge of $T^\text{GT}$ that is being traversed at time step $n$ according to the policy $\Pi$, accumulating the probabilistic weight of that edge if it had not been visited before. The term in the indicator function ensures that each edge is only counted once, the first time that it is visited. The denominator sums the weight of all edges in $T^\text{GT}$, thus returning a normalized weighted coverage. This measure will only reach a value of 1 once all the edges in the ground truth graph $T^\text{GT}$ have been visited. Low-probability edges have a smaller impact in this measure.

Observe that $\operatorname{WCOV}_{T^\text{GT}}(\Pi, N)$ only tells us how good eFeX (or any other algorithm) is at providing an efficient exploration policy $\Pi$ for $T^\text{GT}$, but does not tell us how good the exploration algorithm is in terms of recovering $T^\text{GT}$. In fact, this measure can be used with algorithms that do not have an explicit latent graph that attempts to recover $T^\text{GT}$. Thus, we complement this measure with  another one that focuses on the quality of the model $T$, the \emph{precision}.
    
\paragraph*{Precision} If a model $T$ has recovered a good approximation of the unobservable latent graph $T^\text{GT}$, the hidden states of $T$ should map well to the latent states of $T^\text{GT}$. As discussed in Appendix \ref{sec:app_degenerate}, $T$ might contain several ``split'' hidden states that map to the same latent state of $T^\text{GT}$, and this degenerate solution can still be perfect in terms of modeling $T^\text{GT}$. 
Thus, a way to assess the quality of a model given a sequence of latent ground truth states $\{z_n\}_{n=1}^N$ and the corresponding sequence of  inferred hidden states $\{\hat{z}_n\}_{n=1}^N$ estimated by the model $T$ from observations $\{x_n, a_n\}_{n=1}^N$, is to measure \emph{the accuracy of a predictor that uses the best fixed mapping} from each estimated state $\hat{z}_n$ to each ground truth state $z_n$. Mathematically, we define this precision for the case of an infinitely long walk under a uniform random policy,
\begin{equation}
    \operatorname{PREC}_{T^\text{GT}}(T) = \lim_{N\to\infty}\mathbb{E}_{a_1,\ldots,a_N\sim \text{U}}\Big\{\frac{1}{N} 
    \sum_i \max_j \sum_n \mathbb{1}[\hat{z}_n=i]\mathbb{1}[z_n=j]\Big\}.
    \label{eq:precision}
\end{equation}
Observe that this measure penalizes models that estimate the same hidden state for multiple different ground truth latent states (since each hidden state from the model $T$ can only be mapped to a single latent state from the ground truth $T^\text{GT}$, all the other latent states will register as errors and reduce the accuracy of the predictor). 
Conveniently, this measure does not penalize degenerate solutions: having multiple ``split'' hidden states map to the same ground truth latent state will still not incur in any prediction errors. Thus, perfect degenerate solutions with no stochasticity should have a precision of 1, with smaller values implying that the recovered $T$ is conflating multiple latent states from $T^\text{GT}$ into the same hidden state in $T$. With stochasticity, the precision of a perfect graph might not reach 1, but the measure is still interpretable, since it tells us how well the agent is able to locate itself in the true latent space.

In our experiments, we will approximate these measures by performing $100$ walks starting at random locations of $10,000$ steps each, and computing  $\{\hat{z}_n\}_{n=1}^N$ from the obtained walk data $\{x_n, a_n\}_{n=1}^N$ using Viterbi decoding (when necessary, the expected log-likelihood does not need this step).

\section{Degenerate solutions}
\label{sec:app_degenerate}

Since that for a given ground truth $T^\text{GT}$ perfect recovery results in ${\cal L}_{T^\text{GT}}(T)={\cal L}_{T^\text{GT}}(T^\text{GT})$, one might (incorrectly) assume that ${\cal L}_{T^\text{GT}}(T)={\cal L}_{T^\text{GT}}(T^\text{GT})$ implies that $T=T^\text{GT}$. This is not true, and multiple degenerate solutions with the same expected log-likelihood can be obtained. Two mechanisms are at play in this degeneracy:

\paragraph{Node relabeling} Given a ground truth graph $T^\text{GT}$, an isomorphic graph can be obtained by relabeling its nodes. If $T^\text{GT}$ and $T$ are isomorphic, then ${\cal L}_{T^\text{GT}}(T)={\cal L}_{T^\text{GT}}(T^\text{GT})$. This is the only mechanism for degeneracy in the simple case in which both the original and degenerate graph have no aliasing (i.e., each node emits a unique observation) and a single action is available (i.e., actions are irrelevant). With those conditions, $T^\text{GT}$ and $T$ are isomorphic if and only if ${\cal L}_{T^\text{GT}}(T)={\cal L}_{T^\text{GT}}(T^\text{GT})$.

\paragraph{Clone merging and splitting} In graphs in which aliasing is permitted, it is possible to split a node into two clones of the same node with the same incoming and outgoing edges, and this operation can also be reversed (merging). If only merging and splitting operations are performed on a ground truth graph $T^\text{GT}$, the obtained graph $T$ will have ${\cal L}_{T^\text{GT}}(T)={\cal L}_{T^\text{GT}}(T^\text{GT})$. See Figure \ref{fig:degeneracy_and_TE} for an example.

Since the agent has no way to tell apart degenerate solutions (e.g., an isomorphism) from the actual ground truth, and since a degenerate solution $T$ for which ${\cal L}_{T^\text{GT}}(T)={\cal L}_{T^\text{GT}}(T^\text{GT})$ is, for most practical purposes, just as useful as having access to $T^\text{GT}$, we regard all solutions with the same ${\cal L}_{T^\text{GT}}(T)$ as equally good solutions.

\section{Derivation of Utility}
\label{sec:app_proof}
In this section we provide the derivation for our expression to compute utility:
\begin{equation}
    \nonumber u(z, a) = H\Big(\frac{b_{az}}{1^\top b_{az}}\Big)+ \Big(\frac{1^\top (b_{az}\odot\psi(b_{az}+1))}{1^\top b_{az}}\Big) -\psi(1^\top b_{az} + 1),
\end{equation}    
With begin with Equation~\ref{eq:utility}:
\begin{equation}
u(z, a)  = 
H({\mathbb E}_{t_{az}\sim \operatorname{Dir}(b_{az})}[t_{az}]) - 
{\mathbb E}_{t_{az}\sim \operatorname{Dir}(b_{az})}[H(t_{az})].
\end{equation}

For convenience, we drop the subscript $za$ for this proof and write the utility $u$ of a particular state $z$ and action $a$ pair as:
\begin{equation}
\label{eq:base}
u  = 
H({\mathbb E}_{t\sim \operatorname{Dir}(b)}[t]) - 
{\mathbb E}_{t\sim \operatorname{Dir}(b)}[H(t)],
\end{equation}
where $t$ is the distribution $p(z'|z,a)$ and $b$ is the vector that parameterizes the K+1 dimensional Dirichlet distribution over $t$ (assuming K+1 states $z_{0:K}$). 

The above equation has two components which we analyze separately. The first component is the entropy of the expected transition distribution $t$. Since $t\sim \operatorname{Dir}(b)$, we know that:
\begin{equation}
    {\mathbb E}_{t\sim\operatorname{Dir}(b)}[t] = \frac{b}{\sum_i b[i]},
\end{equation}
which can be alternatively written as
\begin{equation}
\label{eq:term_1_final}
    {\mathbb E}_{t\sim\operatorname{Dir}}(b)[t] = \frac{b}{1^\top b}.
\end{equation}

Now we move to the second term, which is ${\mathbb E}_{t\sim \operatorname{Dir}(b)}[H(t)]$. For a particular sample $t = (t_0, t_1, ..., t_K)$, the entropy $H(t)$ can be written as:
\begin{equation}
    H(t) = -\sum_{i=0}^K t_i \operatorname{log} t_i.
\end{equation}
Then, 
\begin{align}
    {\mathbb E}_{t\sim \operatorname{Dir}(b)}[H(t)] &= -\int_{t\sim\operatorname{Dir}(b)} p(t) \sum_{i=0}^K t_i \operatorname{log} t_i dt\\
    &= -\sum_{i=0}^K \int_{t\sim\operatorname{Dir}(b)} p(t)  t_i \operatorname{log} t_i dt.
\end{align}
We can re-write the previous equation as
\begin{equation}
\label{eq:term_2_as_xi}
    {\mathbb E}_{t\sim \operatorname{Dir}(b)}[H(t)] = -\sum_{i=0}^K x_i
\end{equation}
where $x_i$ is defined as:
\begin{equation}
    \label{eq:xi}
    x_i = \int_{t\sim\operatorname{Dir}(b)} p(t)  t_i \operatorname{log} t_i dt.
\end{equation}
Using the definition of the Dirichlet distribution to expand Equation~\ref{eq:xi} we get:
\begin{equation}
    x_i = \frac{1}{B(b)} \int_{0<=t_{0:K}<=1, \Sigma_i t_i = 1}\Pi_{j} t_j^{b_j-1} t_i \operatorname{log} t_i dt_{0:K},
\end{equation}
where $B(.)$ is the Beta function, and the integral is over the simplex over $t$. To further analyze the above equation, we need to look at the definition of the Beta function:
\begin{equation}
    B(b) = \int_{0<=t_{0:K}<=1, \Sigma_i t_i = 1}\Pi_{j} t_j^{b_j-1} dt_{0:K}.
\end{equation}

Notice the derivative of $B(b)$ w.r.t $b_i$:
\begin{equation}
    \frac{\partial B(b)}{\partial b_i} =  \int_{0<=t_{0:K}<=1, \Sigma_i t_i = 1}\Pi_{j \neq i} t_j^{b_j-1}  t_i^{b_i-1} \operatorname{log} t_i dt_{0:K}.
\end{equation}
If $I_i$ is a one-hot K+1 dimensional vector which is 0 everywhere and 1 at index $i$, then we can write $x_i$ as:
\begin{equation}
    x_i = \frac{1}{B(b)}\frac{\partial B(b+I_i)}{\partial b_i}.
\end{equation}
The derivative of the Beta function is a well known quantity:
\begin{equation}
    \frac{\partial B(b+I_i)}{\partial b_i} = B(b+I_i) \Big( \psi(b_i+1) - \psi(1+\sum_j b_j)\Big),
\end{equation}
where $\psi(.)$ is the polygamma function. Using this expansion we have:
\begin{equation}
\label{eq:xi_gamma}
x_i = \frac{B(b+I_i)}{B(b)} \Big( \psi(b_i+1) - \psi(1+\sum_j b_j)\Big).
\end{equation}
Using the alternative definition of $B(b)$ in terms of the gamma function
\begin{equation}
    B(b) = \frac{1}{\Gamma(\sum_j b_j)}\Pi_i \Gamma(b_i),
\end{equation}
we get:
\begin{equation}
    \frac{B(b+I_i)}{B(b)} = \frac{\Gamma (b_i+1) \Gamma(\sum_j b_j)}{\Gamma (b_i) \Gamma(1+\sum_j b_j)}.
\end{equation}
The gamma function has the property that $\Gamma(x+1) = x\Gamma(x)$, which simplifies the above expression as:
\begin{equation}
    \label{eq:beta_breakdown}
    \frac{B(b+I_i)}{B(b)} = \frac{b_i}{\sum_j b_j}.
\end{equation}
Substituting Equation~\ref{eq:beta_breakdown} in Equation~\ref{eq:xi_gamma}, we have:
\begin{equation}
    \label{eq:xi_final}
    x_i = \frac{B(b+I_i)}{B(b)} = \frac{b_i}{\sum_j b_j} \Big( \psi(b_i+1) - \psi(1+\sum_j b_j)\Big).
\end{equation}
Combining Equations~\ref{eq:xi_final} and~\ref{eq:term_2_as_xi} we have:
\begin{equation}
{\mathbb E}_{t\sim \operatorname{Dir}(b)}[H(t)] = -\sum_{i=0}^K \frac{b_i}{\sum_j b_j} \Big( \psi(b_i+1) - \psi(1+\sum_j b_j)\Big),
\end{equation}
which further simplifies to
\begin{equation}
{\mathbb E}_{t\sim \operatorname{Dir}(b)}[H(t)] = \psi(1+\sum_j b_j)-\sum_{i=0}^K \frac{b_i}{\sum_j b_j} \Big( \psi(b_i+1)\Big). 
\end{equation}
Re-writing the sum as a dot product we get:
\begin{equation}
\label{eq:term_2_final}
{\mathbb E}_{t\sim \operatorname{Dir}(b)}[H(t)] = \psi(1+1^\top b)- 1^\top \Big(\frac{b\odot\psi(b+1)}{1^\top b} \Big).    
\end{equation}

Equation~\ref{eq:term_1_final} and Equation~\ref{eq:term_2_final} can be substituted into Equation~\ref{eq:base} to get
\begin{equation}
    \nonumber u = H\Big(\frac{b}{1^\top b}\Big)+ \Big(\frac{1^\top (b\odot\psi(b+1))}{1^\top b}\Big) -\psi(1^\top b + 1),
\end{equation}
which completes the proof. 

\section{Additional experiments (EM + Viterbi training)}
\label{sec:app_exp}
\newpage

\subsection{Parameters: \texorpdfstring{$\operatorname{UF}=1.0$ and $P_\text{slip}=0.00$}{UF = 1.0 and pSLIP = 0.00}}
\begin{figure}[!htb]
    \centering
    \subfigure{\includegraphics[width=0.9\textwidth]{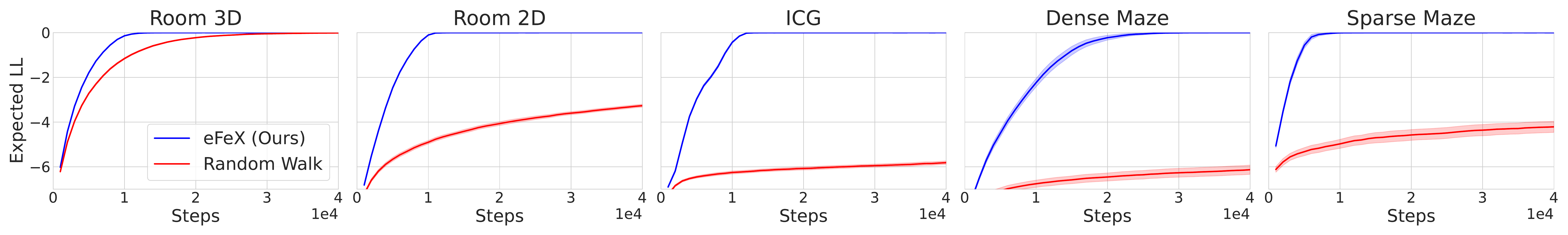}}
    \subfigure{\includegraphics[width=0.9\textwidth]{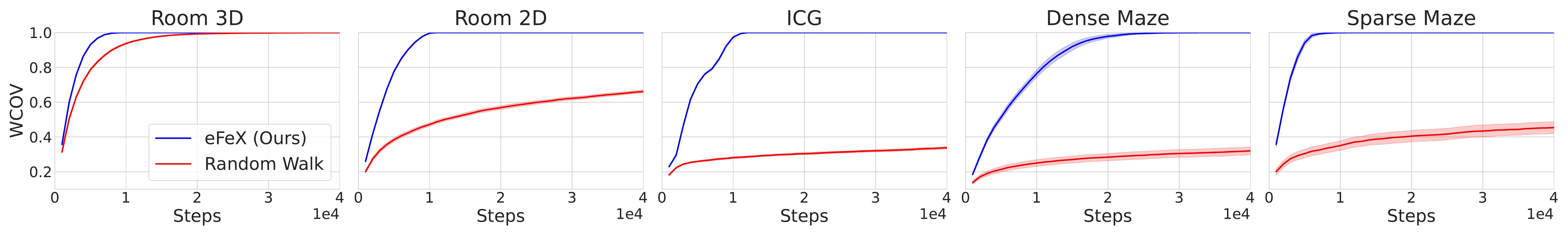}}
    \subfigure{\includegraphics[width=0.9\textwidth]{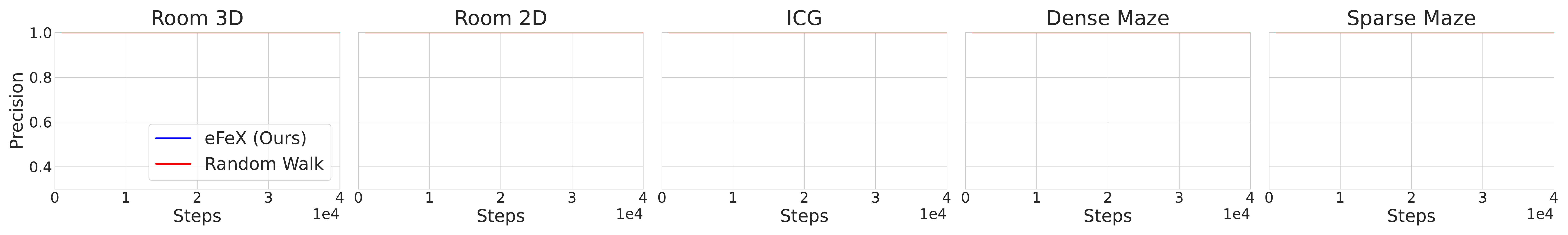}}
    \caption{Expected log-likelihood, weighted coverage, and precision, for each topology, as a function of the number of steps of the agent in the environment. See Appendix \ref{sec:app_performance} for further explanation. Higher is better. Results are averaged over 40 runs. Bands indicate 95\% confidence intervals for the averages. Random policy shown in blue, eFeX in red.}
\end{figure}

\subsection{Parameters: \texorpdfstring{$\operatorname{UF}=1.0$ and $P_\text{slip}=0.01$}{UF = 1.0 and pSLIP = 0.01}}
\begin{figure}[!htb]
    \centering
    \subfigure{\includegraphics[width=0.9\textwidth]{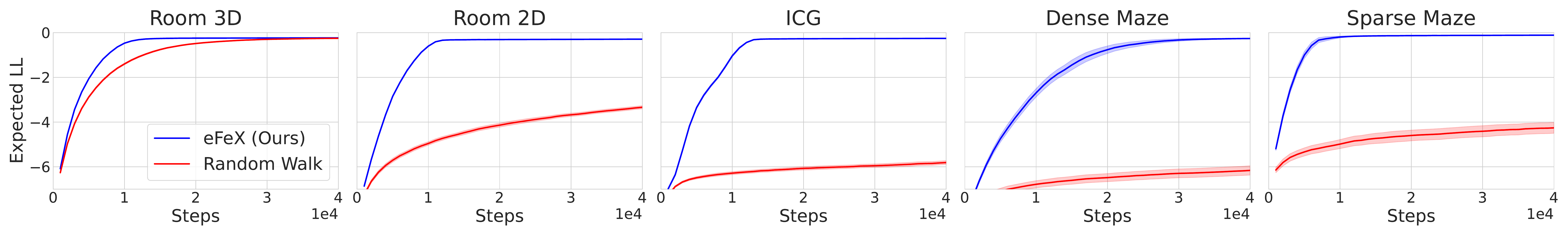}}
    \subfigure{\includegraphics[width=0.9\textwidth]{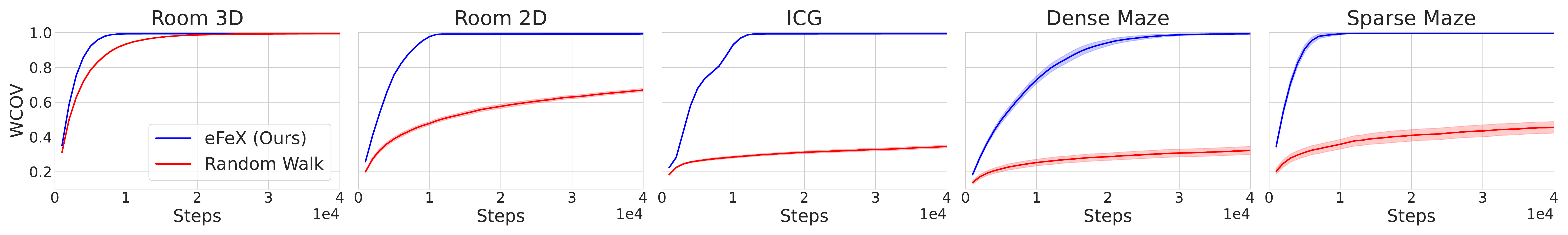}}
    \subfigure{\includegraphics[width=0.9\textwidth]{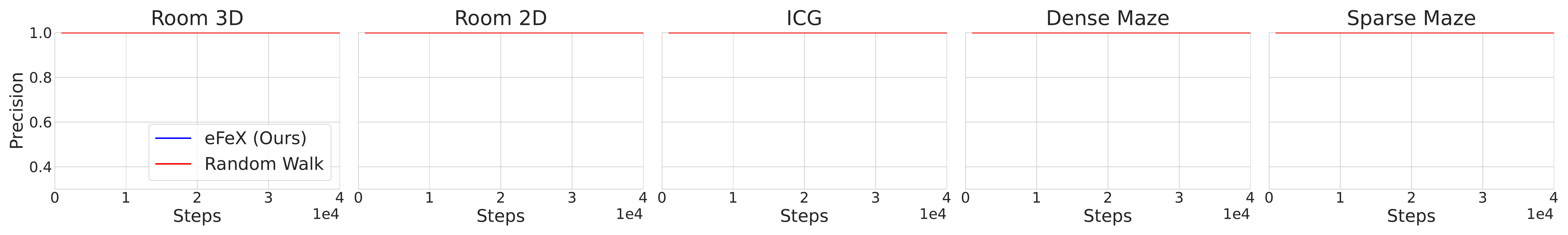}}
    \caption{Expected log-likelihood, weighted coverage, and precision, for each topology, as a function of the number of steps of the agent in the environment. See Appendix \ref{sec:app_performance} for further explanation. Higher is better. Results are averaged over 40 runs. Bands indicate 95\% confidence intervals for the averages. Random policy shown in blue, eFeX in red.}
\end{figure}

\newpage

\subsection{Parameters: \texorpdfstring{$\operatorname{UF}=1.0$ and $P_\text{slip}=0.10$}{UF = 1.0 and pSLIP = 0.10}}
\begin{figure}[!htb]
    \centering
    \subfigure{\includegraphics[width=0.9\textwidth]{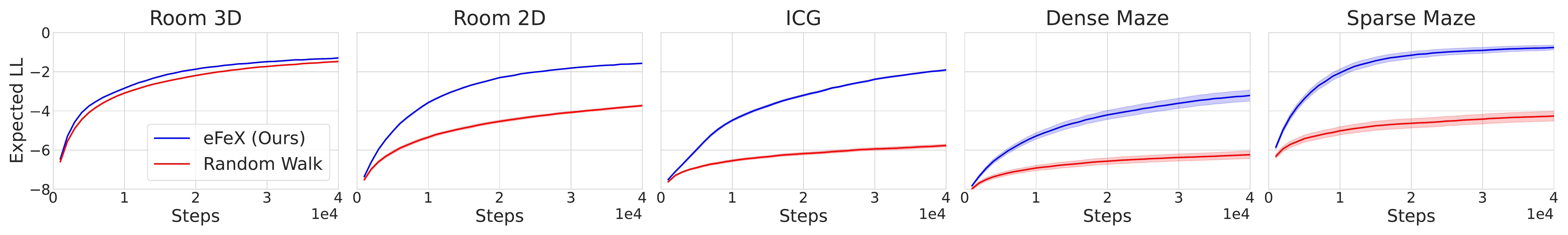}}
    \subfigure{\includegraphics[width=0.9\textwidth]{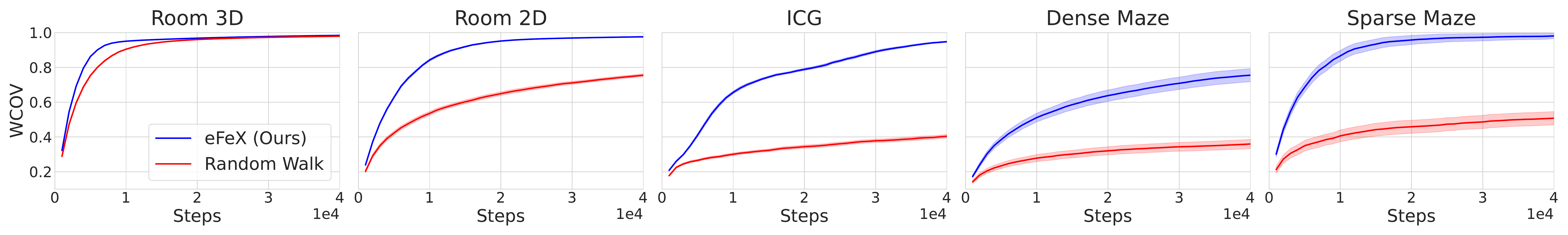}}
    \subfigure{\includegraphics[width=0.9\textwidth]{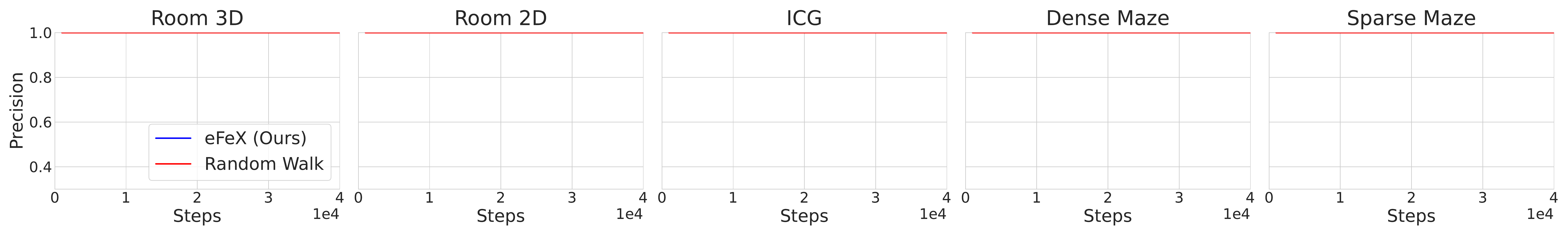}}
    \caption{Expected log-likelihood, weighted coverage, and precision, for each topology, as a function of the number of steps of the agent in the environment. See Appendix \ref{sec:app_performance} for further explanation. Higher is better. Results are averaged over 40 runs. Bands indicate 95\% confidence intervals for the averages. Random policy shown in blue, eFeX in red.}
\end{figure}

\subsection{Parameters: \texorpdfstring{$\operatorname{UF}=0.3$ and $P_\text{slip}=0.00$}{UF = 0.3 and pSLIP = 0.00}}
\begin{figure}[!htb]
    \centering
    \subfigure{\includegraphics[width=0.9\textwidth]{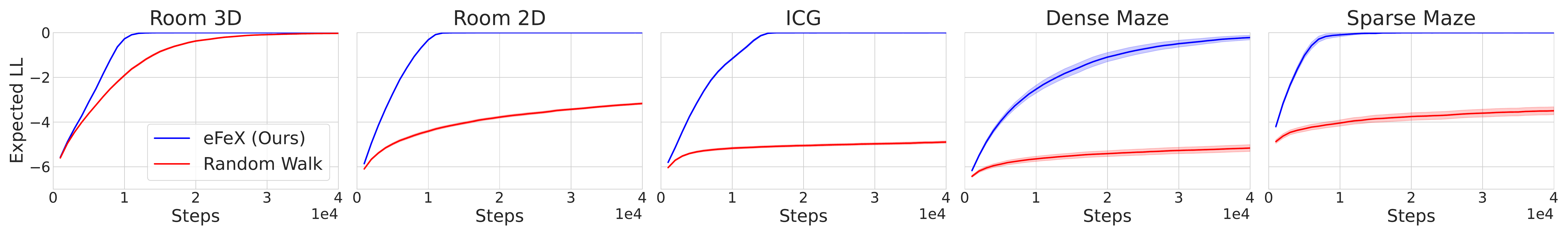}}
    \subfigure{\includegraphics[width=0.9\textwidth]{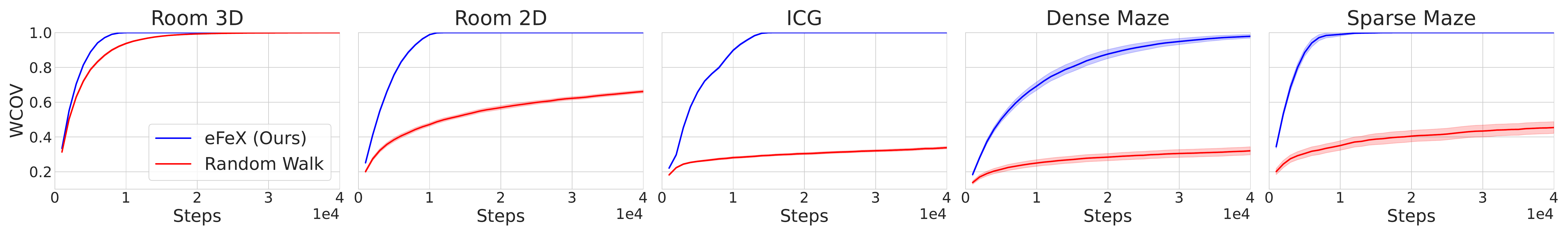}}
    \subfigure{\includegraphics[width=0.9\textwidth]{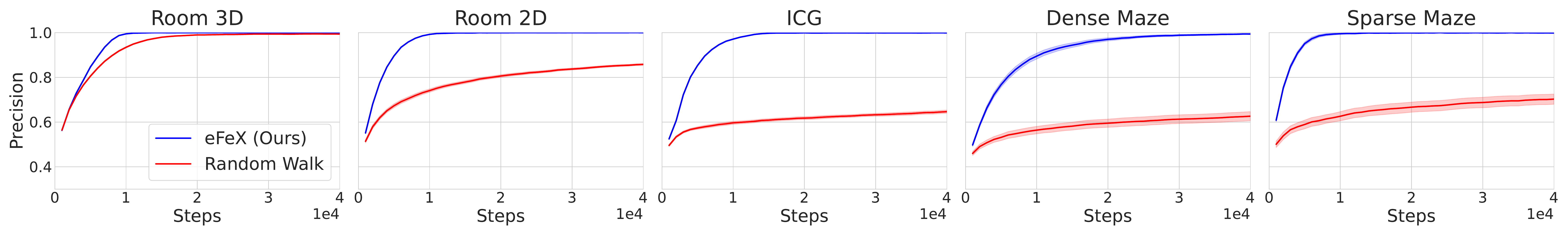}}
    \caption{Expected log-likelihood, weighted coverage, and precision, for each topology, as a function of the number of steps of the agent in the environment. See Appendix \ref{sec:app_performance} for further explanation. Higher is better. Results are averaged over 40 runs. Bands indicate 95\% confidence intervals for the averages. Random policy shown in blue, eFeX in red.}
\end{figure}


\subsection{Parameters: \texorpdfstring{$\operatorname{UF}=0.3$ and $P_\text{slip}=0.01$}{UF = 0.3 and pSLIP = 0.01}}
\begin{figure}[!htb]
    \centering
    \subfigure{\includegraphics[width=0.9\textwidth]{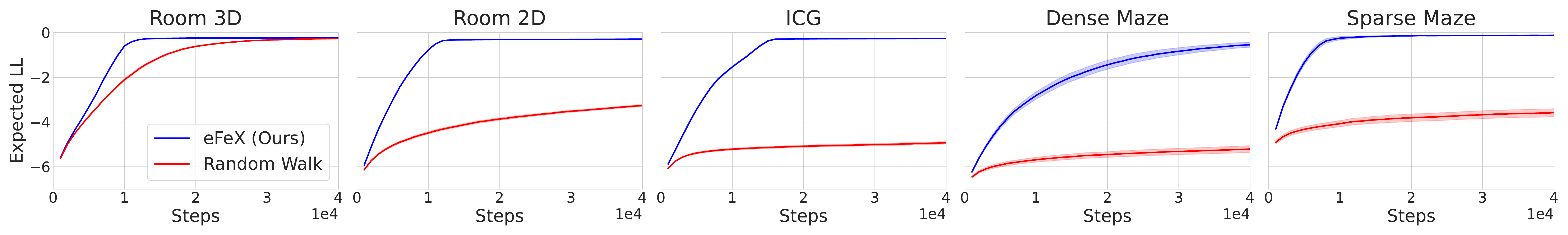}}
    \subfigure{\includegraphics[width=0.9\textwidth]{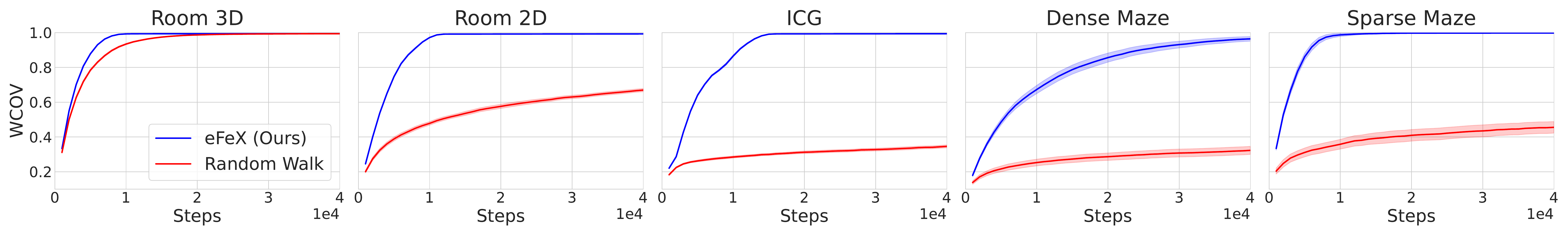}}
    \subfigure{\includegraphics[width=0.9\textwidth]{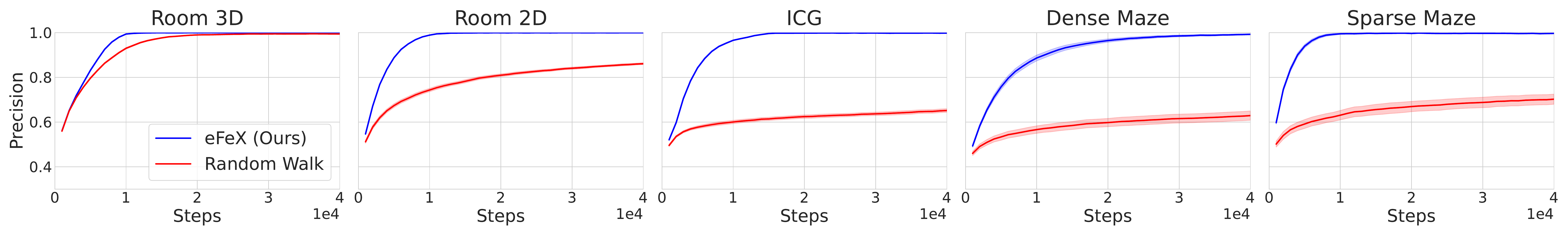}}
    \caption{Expected log-likelihood, weighted coverage, and precision, for each topology, as a function of the number of steps of the agent in the environment. See Appendix \ref{sec:app_performance} for further explanation. Higher is better. Results are averaged over 40 runs. Bands indicate 95\% confidence intervals for the averages. Random policy shown in blue, eFeX in red.}
\end{figure}

\subsection{Parameters: \texorpdfstring{$\operatorname{UF}=0.3$ and $P_\text{slip}=0.10$}{UF = 0.3 and pSLIP = 0.10}}
\begin{figure}[!htb]
    \centering
    \subfigure{\includegraphics[width=0.9\textwidth]{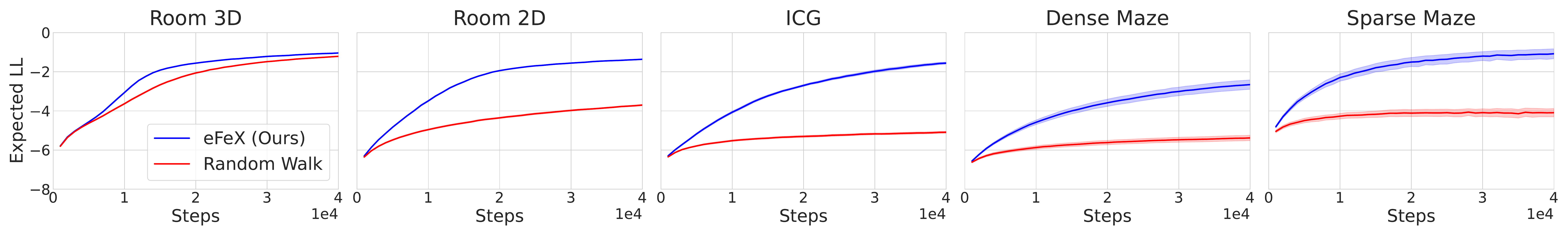}}
    \subfigure{\includegraphics[width=0.9\textwidth]{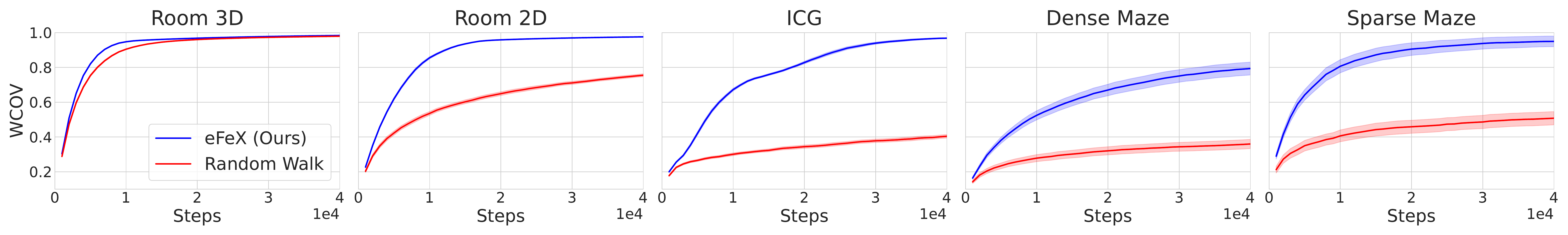}}
    \subfigure{\includegraphics[width=0.9\textwidth]{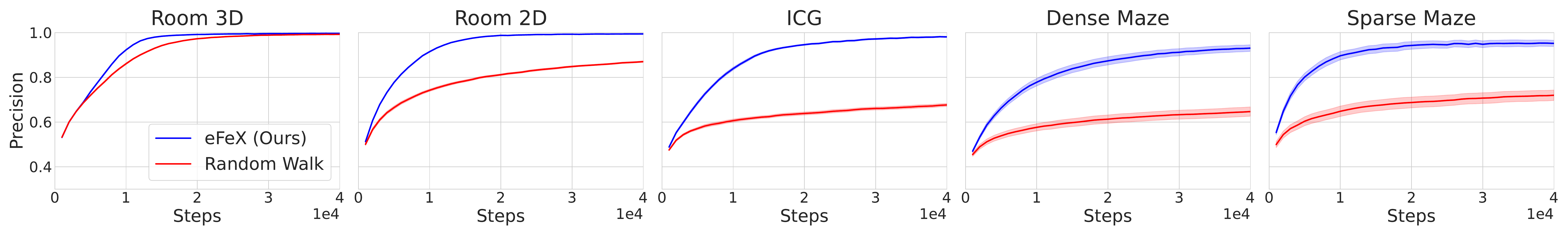}}
    \caption{Expected log-likelihood, weighted coverage, and precision, for each topology, as a function of the number of steps of the agent in the environment. See Appendix \ref{sec:app_performance} for further explanation. Higher is better. Results are averaged over 40 runs. Bands indicate 95\% confidence intervals for the averages. Random policy shown in blue, eFeX in red.}
\end{figure}


\subsection{Parameters: \texorpdfstring{$\operatorname{UF}=0.1$ and $P_\text{slip}=0.00$}{UF = 0.1 and pSLIP = 0.00}}
\begin{figure}[!htb]
    \centering
    \subfigure{\includegraphics[width=0.9\textwidth]{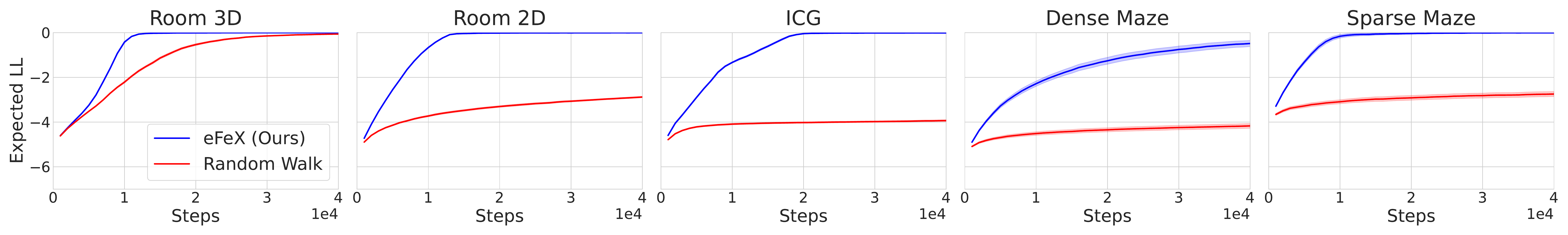}}
    \subfigure{\includegraphics[width=0.9\textwidth]{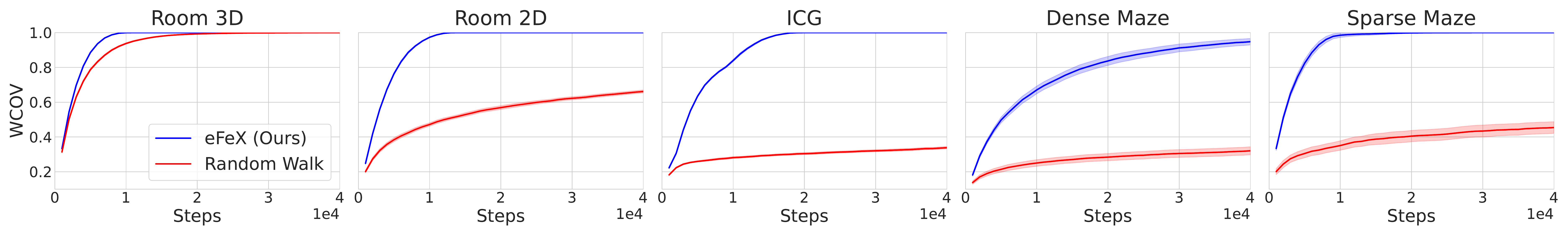}}
    \subfigure{\includegraphics[width=0.9\textwidth]{figs/100_40/precision_uf_0.1_sp_0.0.pdf}}
    \caption{Expected log-likelihood, weighted coverage, and precision, for each topology, as a function of the number of steps of the agent in the environment. See Appendix \ref{sec:app_performance} for further explanation. Higher is better. Results are averaged over 40 runs. Bands indicate 95\% confidence intervals for the averages. Random policy shown in blue, eFeX in red.}
\end{figure}

\subsection{Parameters: \texorpdfstring{$\operatorname{UF}=0.1$ and $P_\text{slip}=0.01$}{UF = 0.1 and pSLIP = 0.01}}
\begin{figure}[!htb]
    \centering
    \subfigure{\includegraphics[width=0.9\textwidth]{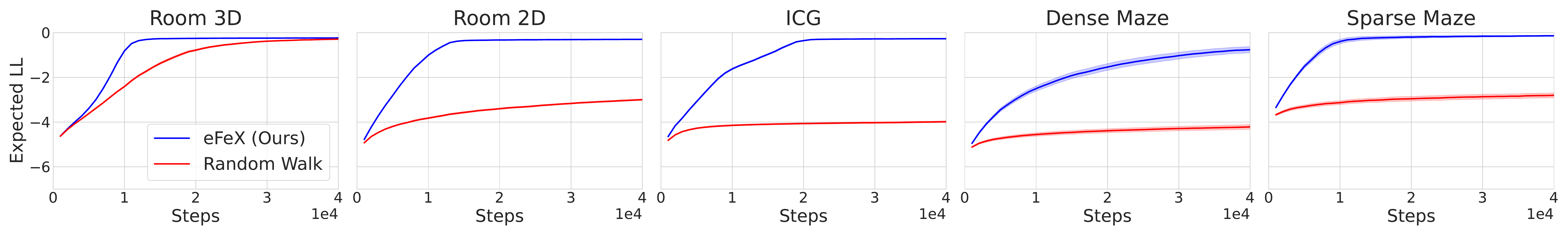}}
    \subfigure{\includegraphics[width=0.9\textwidth]{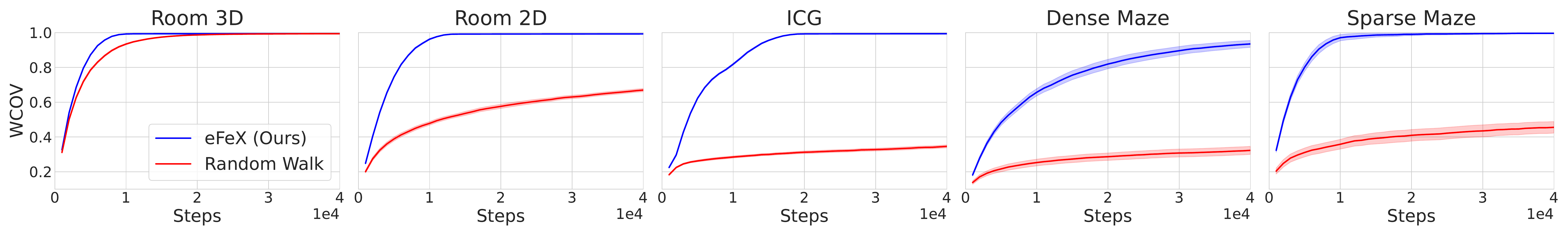}}
    \subfigure{\includegraphics[width=0.9\textwidth]{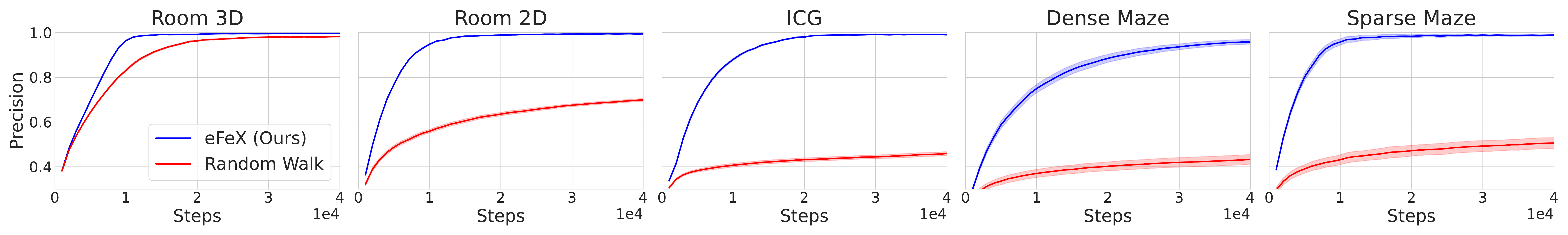}}
    \caption{Expected log-likelihood, weighted coverage, and precision, for each topology, as a function of the number of steps of the agent in the environment. See Appendix \ref{sec:app_performance} for further explanation. Higher is better. Results are averaged over 40 runs. Bands indicate 95\% confidence intervals for the averages. Random policy shown in blue, eFeX in red.}
\end{figure}


\subsection{Parameters: \texorpdfstring{$\operatorname{UF}=0.1$ and $P_\text{slip}=0.10$}{UF = 0.1 and pSLIP = 0.10}}
\begin{figure}[!htb]
    \centering
    \subfigure{\includegraphics[width=0.9\textwidth]{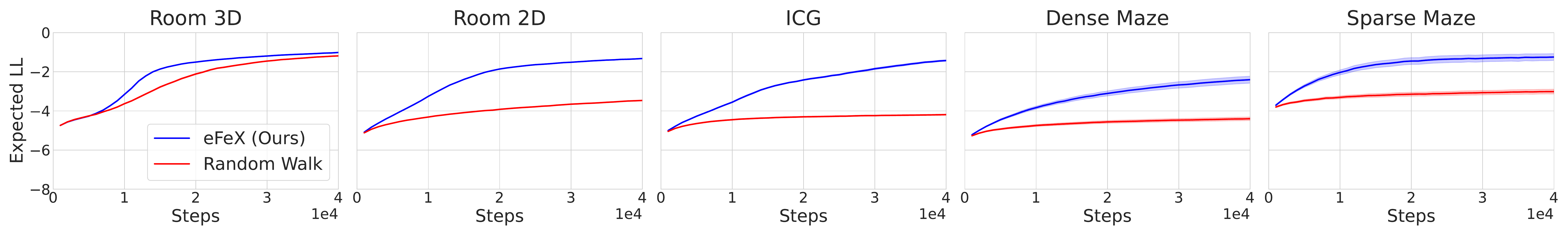}}
    \subfigure{\includegraphics[width=0.9\textwidth]{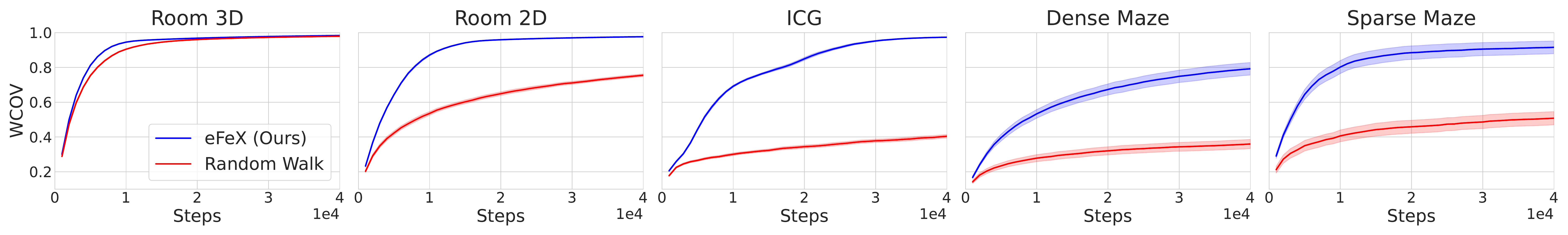}}
    \subfigure{\includegraphics[width=0.9\textwidth]{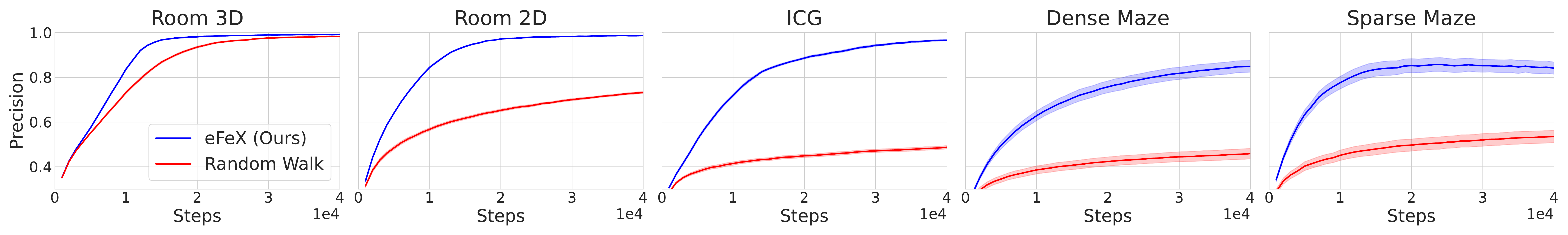}}
    \caption{Expected log-likelihood, weighted coverage, and precision, for each topology, as a function of the number of steps of the agent in the environment. See Appendix \ref{sec:app_performance} for further explanation. Higher is better. Results are averaged over 40 runs. Bands indicate 95\% confidence intervals for the averages. Random policy shown in blue, eFeX in red.}
\end{figure}

\section{Performance on aliased graphs with the “maze” topology from Section \ref{sec:aliased_many_topologies}}

\begin{figure}[htb]
\centering
\includegraphics[width=0.5\linewidth]{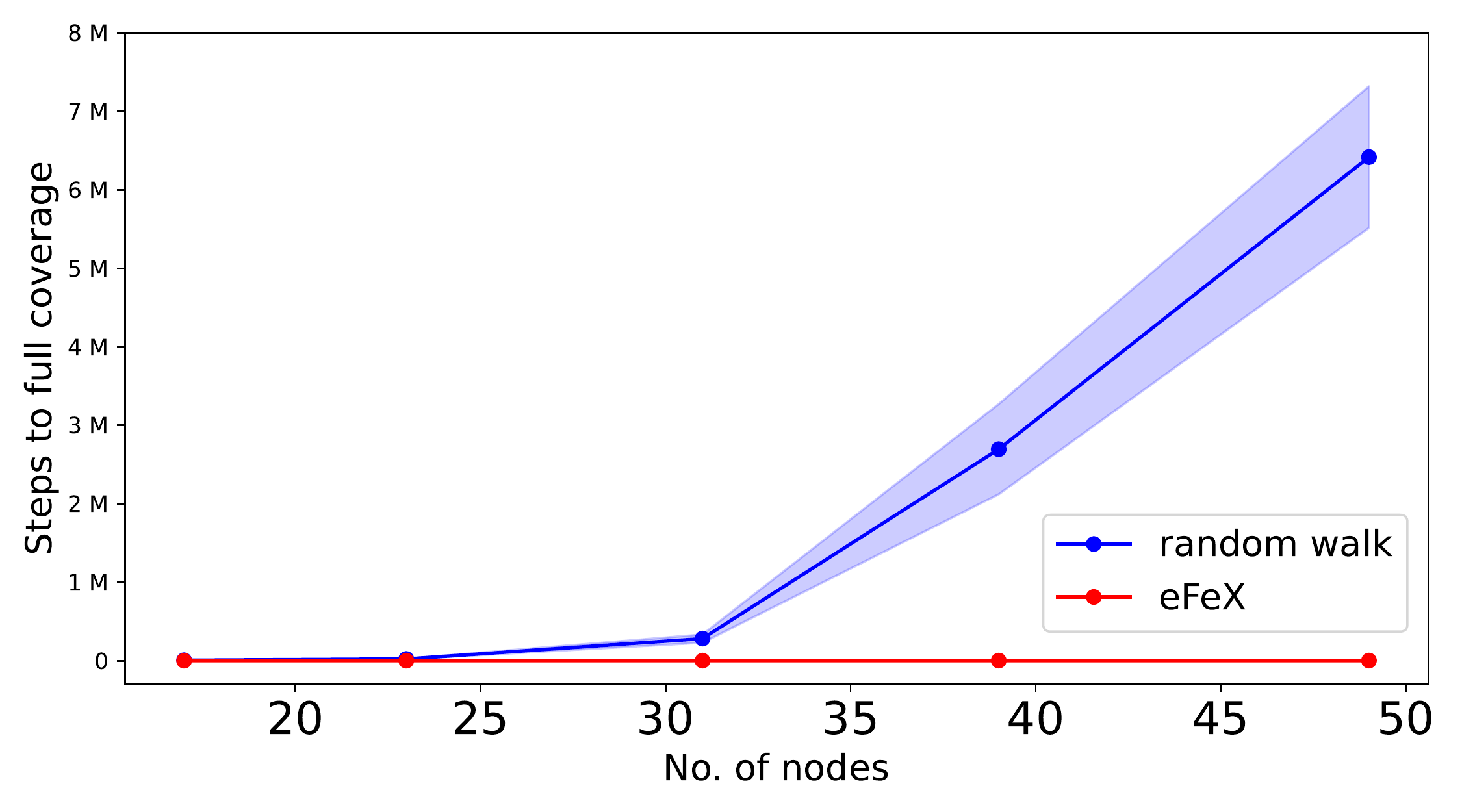}
\caption{Scaling with the size of the ``maze'' topology, under the random and eFeX policies. eFeX is exponentially more efficient.  Averaged over 100 runs. 95\% confidence intervals provided. Same data as Fig.~\ref{fig:chain1d_scaling}[right], but with the Y-axis on a linear scale.}
\label{fig:env_with_resets_log}
\end{figure}

\section{Additional experiments (EM + variational Bayes)}
\label{sec:app_expVB}
See \cite[][Chapter 3]{beal2003variational} for details on how to use variational Bayes (VB) for HMM learning (analogous for CSCG learning). We run EM and then refine and sparsify $T$ by running VB from the EM solution.

\newpage

\subsection{Parameters: \texorpdfstring{$\operatorname{UF}=1.0$ and $P_\text{slip}=0.00$}{UF = 1.0 and pSLIP = 0.00}}
\begin{figure}[!h]
    \centering
    \subfigure{\includegraphics[width=0.9\textwidth]{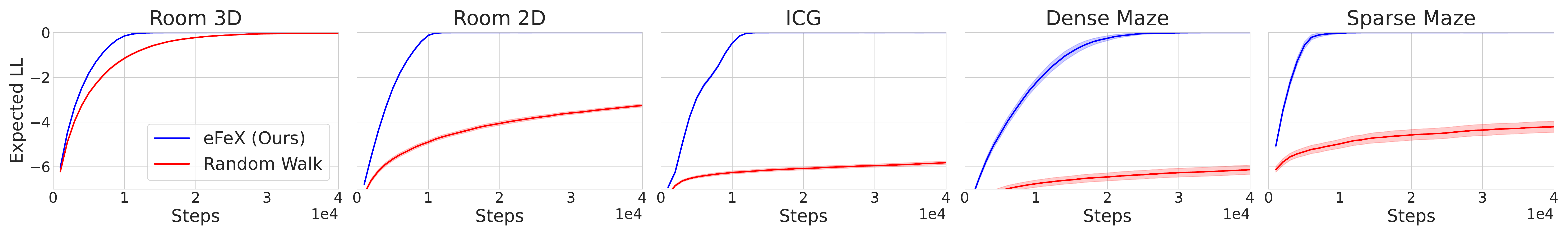}}
    \subfigure{\includegraphics[width=0.9\textwidth]{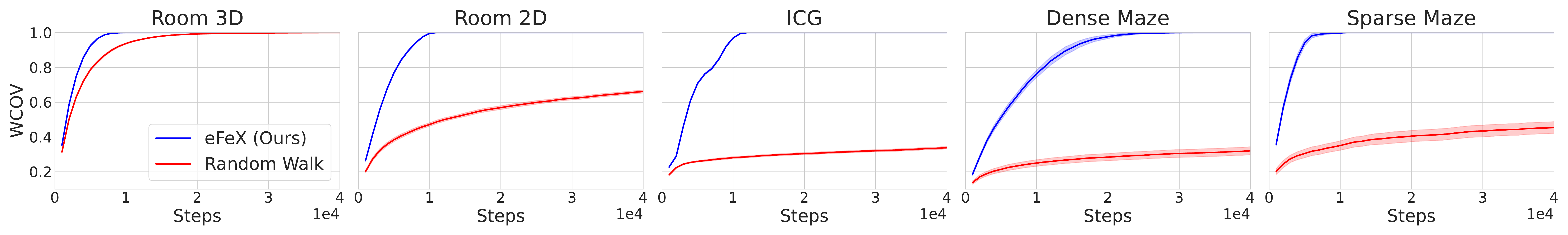}}
    \subfigure{\includegraphics[width=0.9\textwidth]{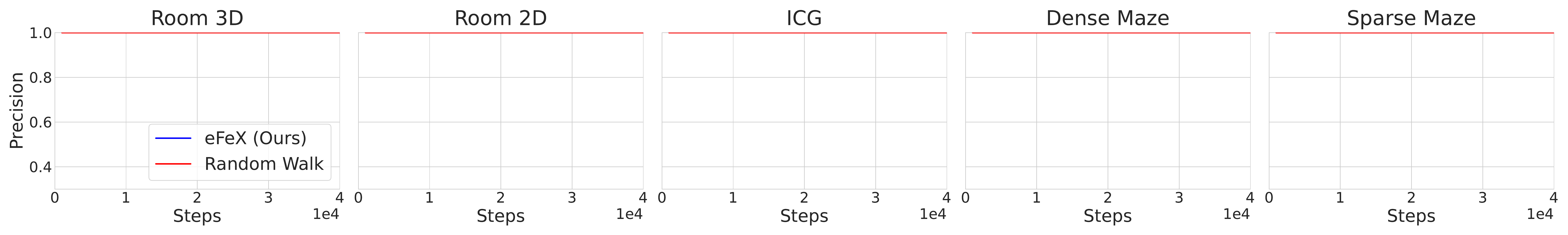}}
    \caption{Expected log-likelihood, weighted coverage, and precision, for each topology, as a function of the number of steps of the agent in the environment. See Appendix \ref{sec:app_performance} for further explanation. Higher is better. Results are averaged over 40 runs. Bands indicate 95\% confidence intervals for the averages. Random policy shown in blue, eFeX in red.}
\end{figure}

\subsection{Parameters: \texorpdfstring{$\operatorname{UF}=1.0$ and $P_\text{slip}=0.01$}{UF = 1.0 and pSLIP = 0.01}}
\begin{figure}[!h]
    \centering
    \subfigure{\includegraphics[width=0.9\textwidth]{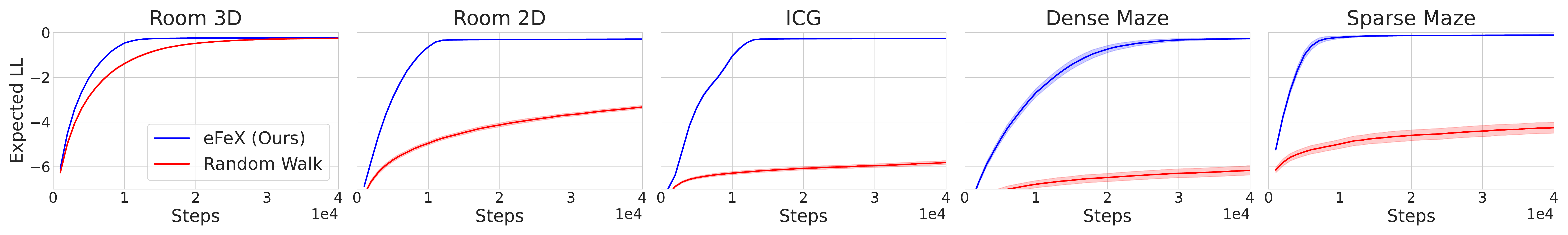}}
    \subfigure{\includegraphics[width=0.9\textwidth]{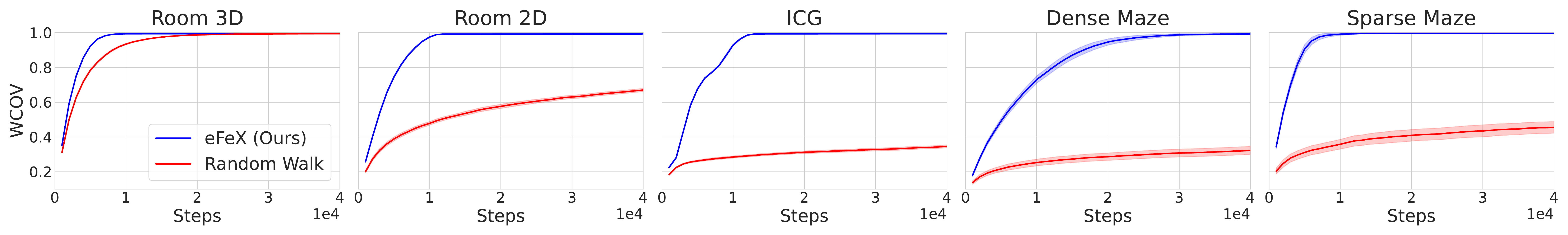}}
    \subfigure{\includegraphics[width=0.9\textwidth]{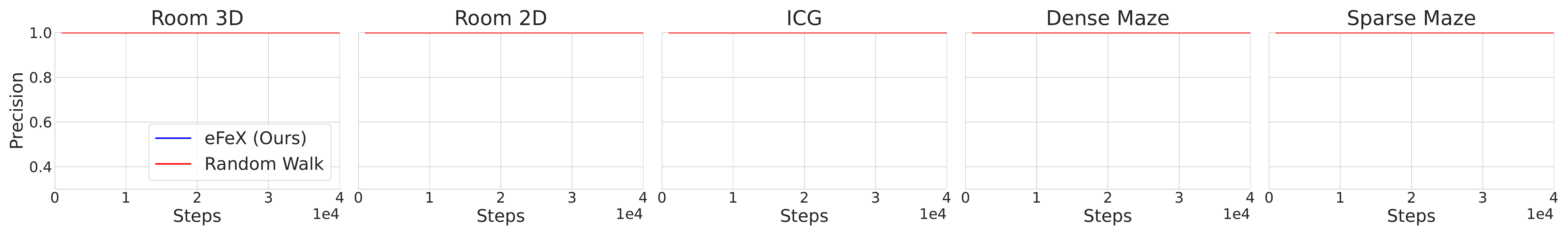}}
    \caption{Expected log-likelihood, weighted coverage, and precision, for each topology, as a function of the number of steps of the agent in the environment. See Appendix \ref{sec:app_performance} for further explanation. Higher is better. Results are averaged over 40 runs. Bands indicate 95\% confidence intervals for the averages. Random policy shown in blue, eFeX in red.}
\end{figure}

\newpage

\subsection{Parameters: \texorpdfstring{$\operatorname{UF}=1.0$ and $P_\text{slip}=0.10$}{UF = 1.0 and pSLIP = 0.10}}
\begin{figure}[!h]
    \centering
    \subfigure{\includegraphics[width=0.9\textwidth]{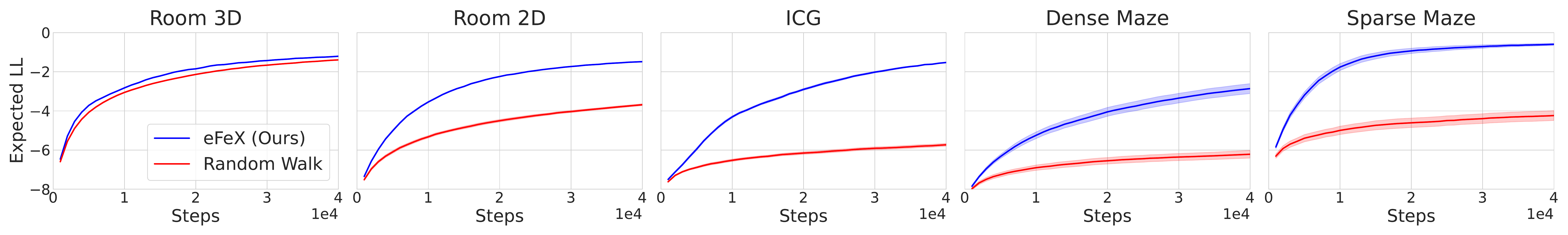}}
    \subfigure{\includegraphics[width=0.9\textwidth]{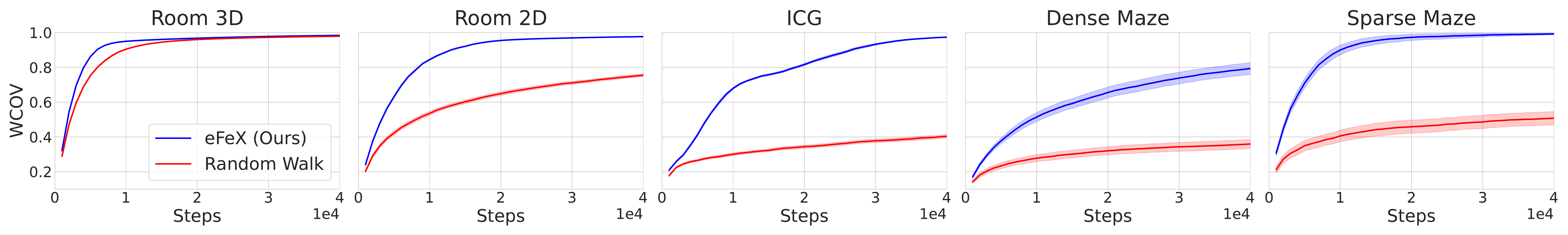}}
    \subfigure{\includegraphics[width=0.9\textwidth]{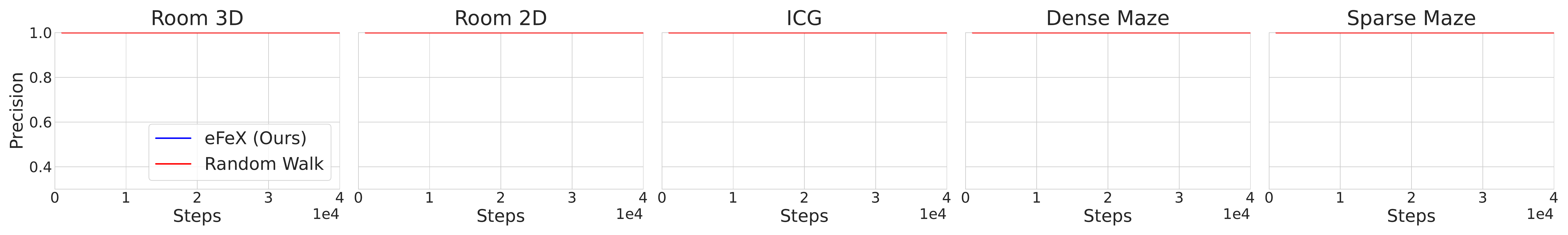}}
    \caption{Expected log-likelihood, weighted coverage, and precision, for each topology, as a function of the number of steps of the agent in the environment. See Appendix \ref{sec:app_performance} for further explanation. Higher is better. Results are averaged over 40 runs. Bands indicate 95\% confidence intervals for the averages. Random policy shown in blue, eFeX in red.}
\end{figure}

\subsection{Parameters: \texorpdfstring{$\operatorname{UF}=0.3$ and $P_\text{slip}=0.00$}{UF = 0.3 and pSLIP = 0.00}}
\begin{figure}[!h]
    \centering
    \subfigure{\includegraphics[width=0.9\textwidth]{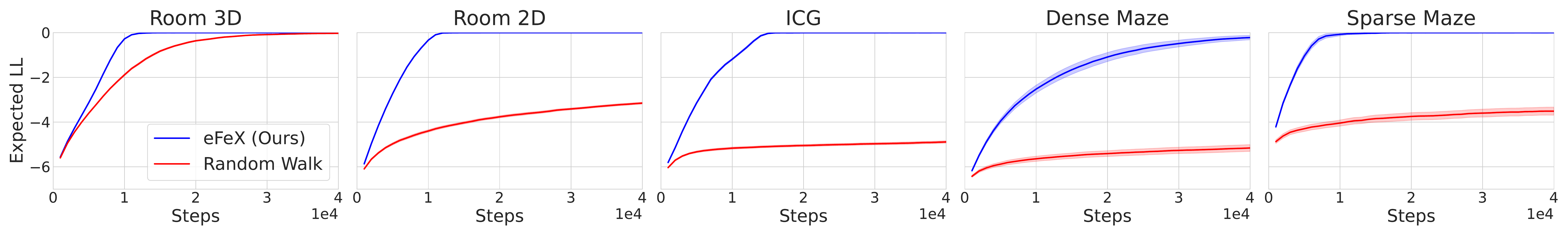}}
    \subfigure{\includegraphics[width=0.9\textwidth]{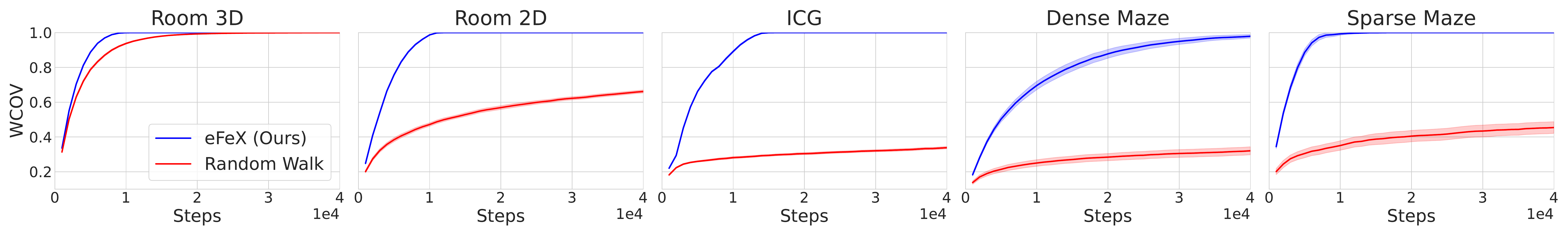}}
    \subfigure{\includegraphics[width=0.9\textwidth]{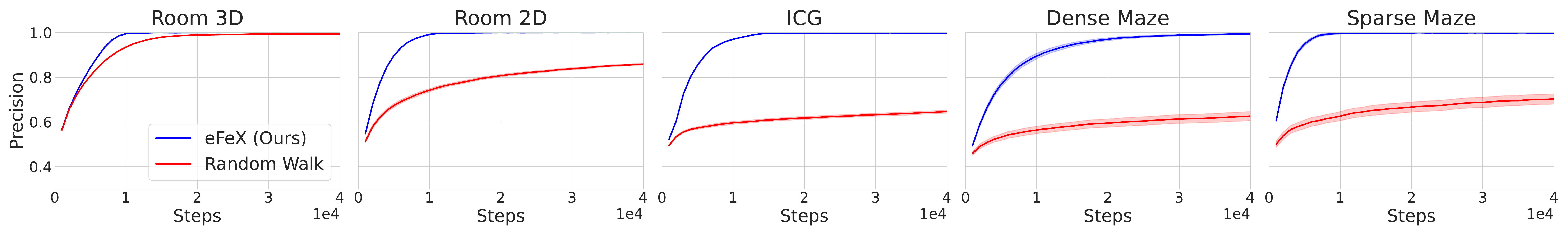}}
    \caption{Expected log-likelihood, weighted coverage, and precision, for each topology, as a function of the number of steps of the agent in the environment. See Appendix \ref{sec:app_performance} for further explanation. Higher is better. Results are averaged over 40 runs. Bands indicate 95\% confidence intervals for the averages. Random policy shown in blue, eFeX in red.}
\end{figure}

\newpage

\subsection{Parameters: \texorpdfstring{$\operatorname{UF}=0.3$ and $P_\text{slip}=0.01$}{UF = 0.3 and pSLIP = 0.01}}
\begin{figure}[!h]
    \centering
    \subfigure{\includegraphics[width=0.9\textwidth]{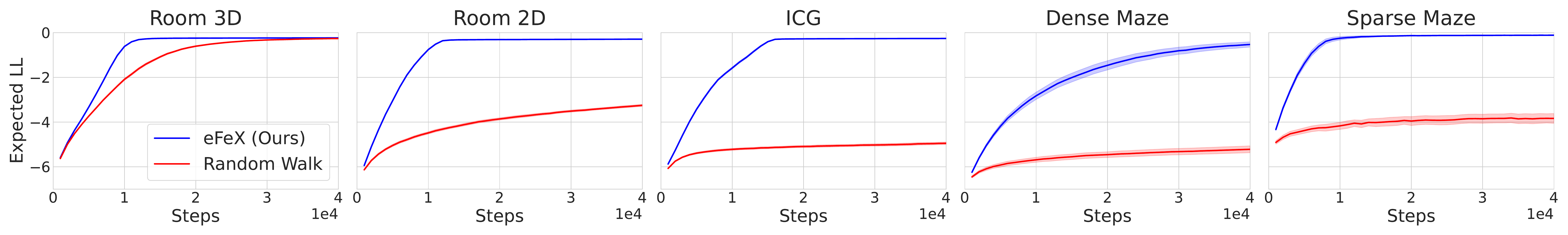}}
    \subfigure{\includegraphics[width=0.9\textwidth]{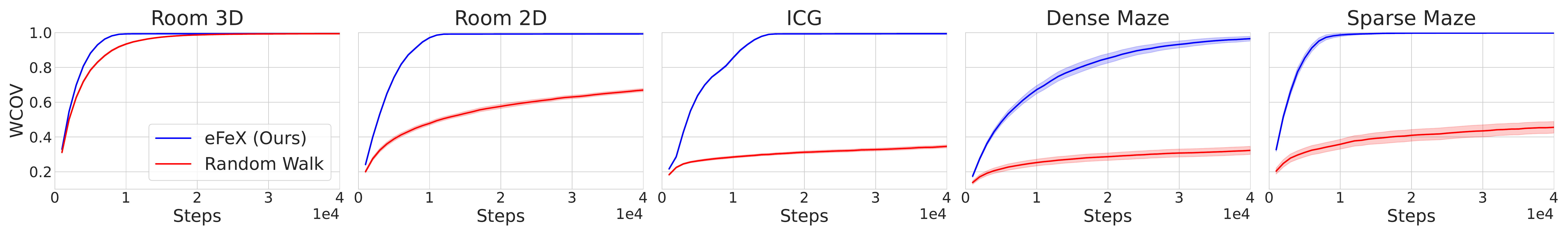}}
    \subfigure{\includegraphics[width=0.9\textwidth]{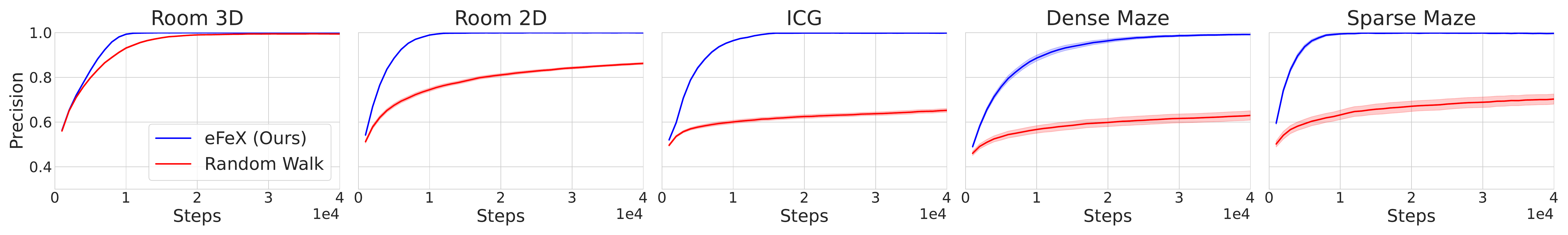}}
    \caption{Expected log-likelihood, weighted coverage, and precision, for each topology, as a function of the number of steps of the agent in the environment. See Appendix \ref{sec:app_performance} for further explanation. Higher is better. Results are averaged over 40 runs. Bands indicate 95\% confidence intervals for the averages. Random policy shown in blue, eFeX in red.}
\end{figure}

\subsection{Parameters: \texorpdfstring{$\operatorname{UF}=0.3$ and $P_\text{slip}=0.10$}{UF = 0.3 and pSLIP = 0.10}}
\begin{figure}[!h]
    \centering
    \subfigure{\includegraphics[width=0.9\textwidth]{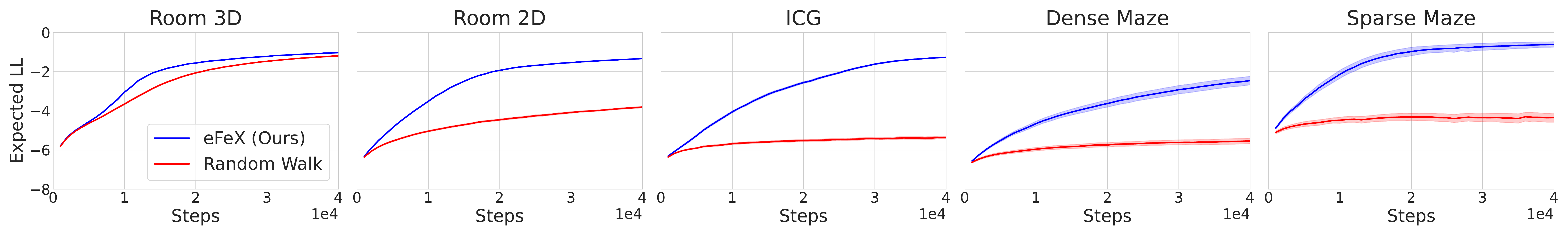}}
    \subfigure{\includegraphics[width=0.9\textwidth]{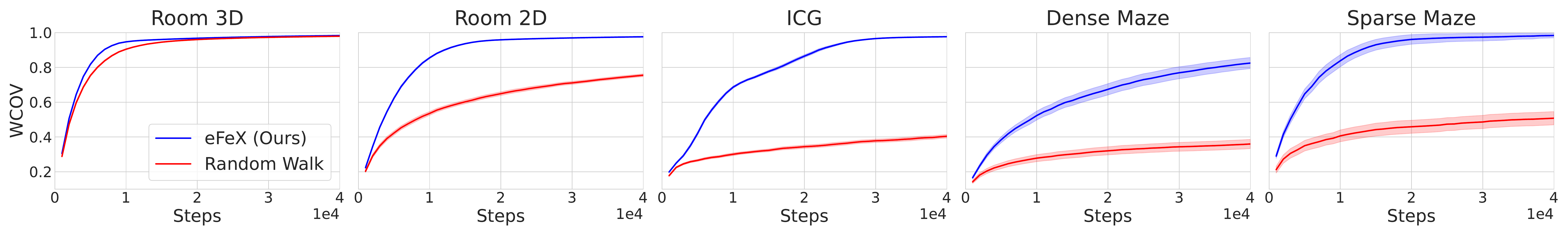}}
    \subfigure{\includegraphics[width=0.9\textwidth]{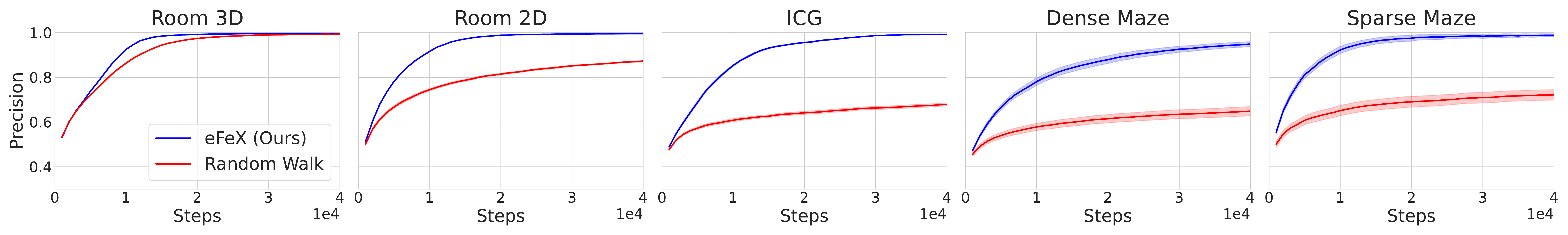}}
    \caption{Expected log-likelihood, weighted coverage, and precision, for each topology, as a function of the number of steps of the agent in the environment. See Appendix \ref{sec:app_performance} for further explanation. Higher is better. Results are averaged over 40 runs. Bands indicate 95\% confidence intervals for the averages. Random policy shown in blue, eFeX in red.}
\end{figure}

\newpage

\subsection{Parameters: \texorpdfstring{$\operatorname{UF}=0.1$ and $P_\text{slip}=0.00$}{UF = 0.1 and pSLIP = 0.00}}
\begin{figure}[!h]
    \centering
    \subfigure{\includegraphics[width=0.9\textwidth]{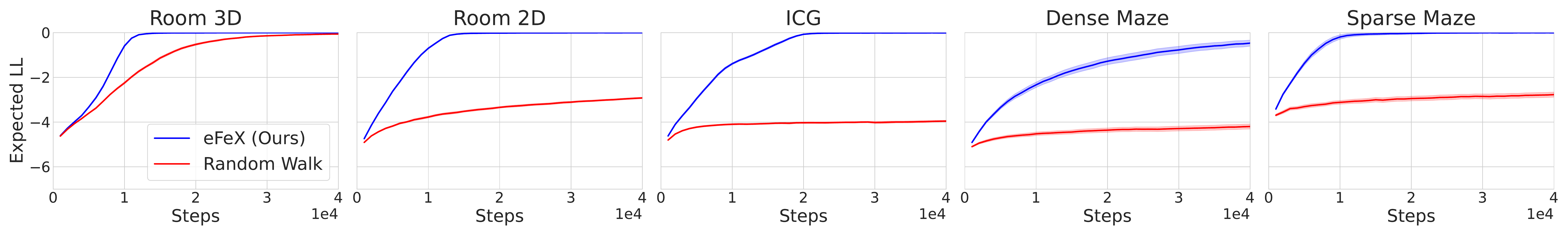}}
    \subfigure{\includegraphics[width=0.9\textwidth]{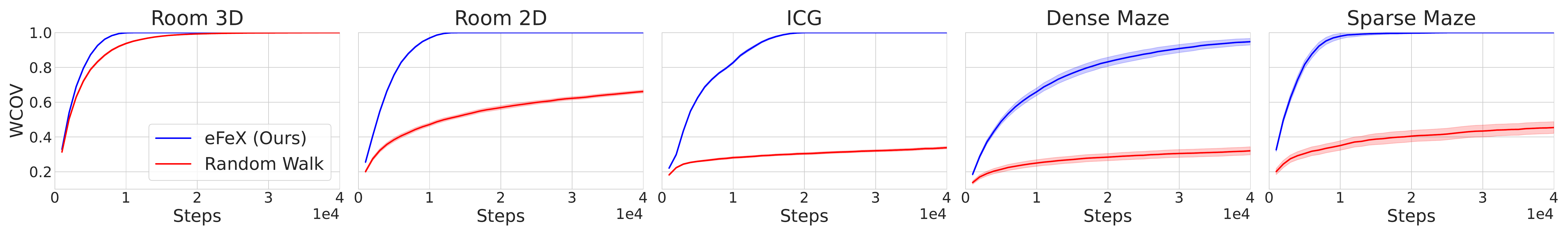}}
    \subfigure{\includegraphics[width=0.9\textwidth]{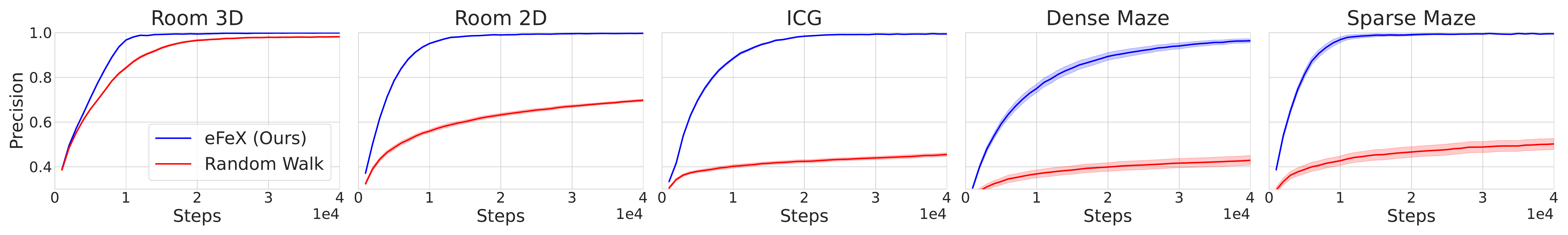}}
    \caption{Expected log-likelihood, weighted coverage, and precision, for each topology, as a function of the number of steps of the agent in the environment. See Appendix \ref{sec:app_performance} for further explanation. Higher is better. Results are averaged over 40 runs. Bands indicate 95\% confidence intervals for the averages. Random policy shown in blue, eFeX in red.}
\end{figure}

\subsection{Parameters: \texorpdfstring{$\operatorname{UF}=0.1$ and $P_\text{slip}=0.01$}{UF = 0.1 and pSLIP = 0.01}}
\begin{figure}[!h]
    \centering
    \subfigure{\includegraphics[width=0.9\textwidth]{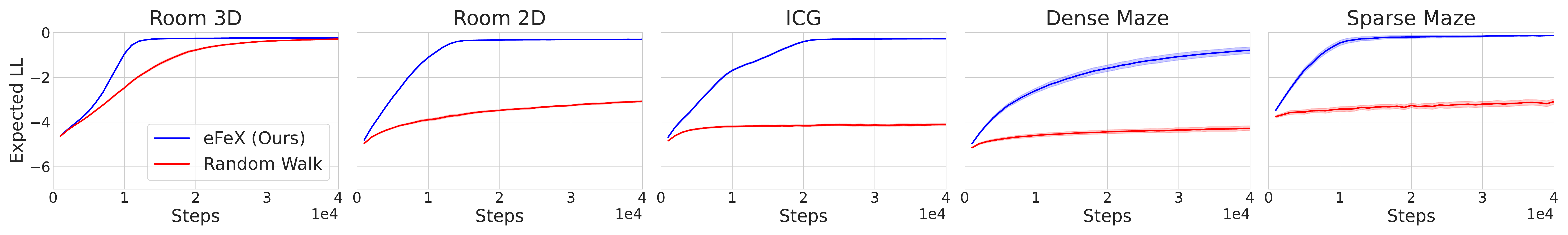}}
    \subfigure{\includegraphics[width=0.9\textwidth]{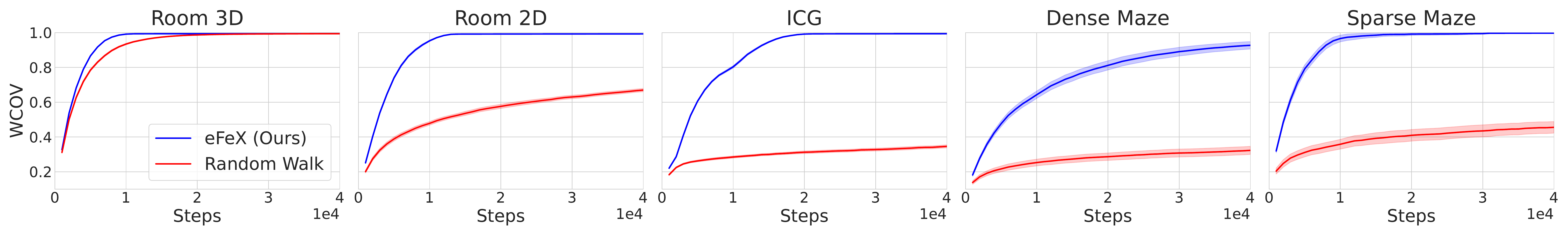}}
    \subfigure{\includegraphics[width=0.9\textwidth]{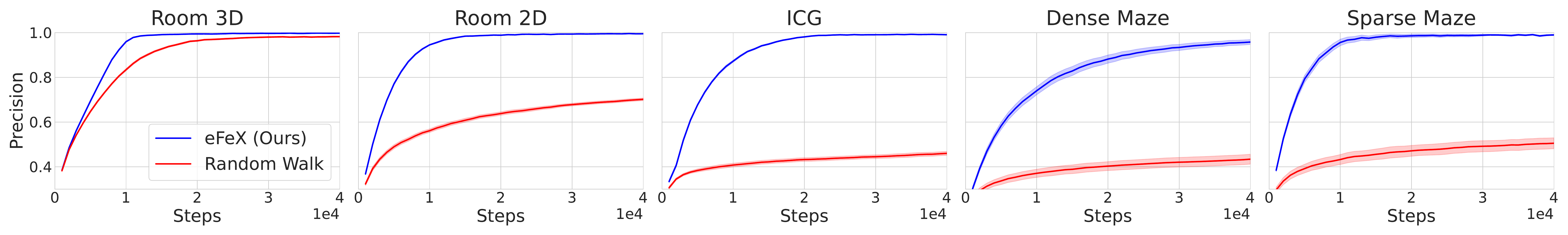}}
    \caption{Expected log-likelihood, weighted coverage, and precision, for each topology, as a function of the number of steps of the agent in the environment. See Appendix \ref{sec:app_performance} for further explanation. Higher is better. Results are averaged over 40 runs. Bands indicate 95\% confidence intervals for the averages. Random policy shown in blue, eFeX in red.}
\end{figure}

\newpage

\subsection{Parameters: \texorpdfstring{$\operatorname{UF}=0.1$ and $P_\text{slip}=0.10$}{UF = 0.1 and pSLIP = 0.10}}
\begin{figure}[!htb]
    \centering
    \subfigure{\includegraphics[width=0.9\textwidth]{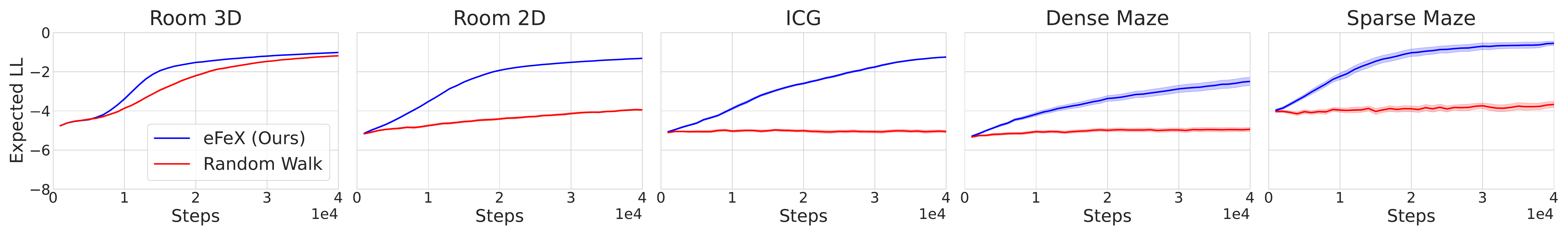}}
    \subfigure{\includegraphics[width=0.9\textwidth]{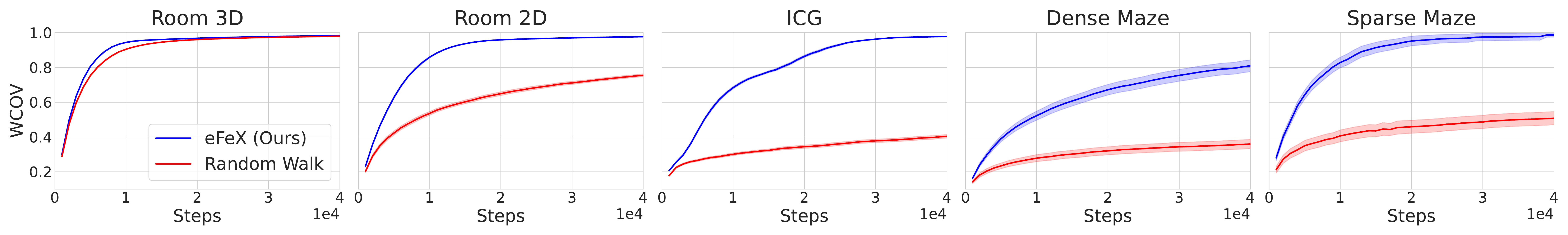}}
    \subfigure{\includegraphics[width=0.9\textwidth]{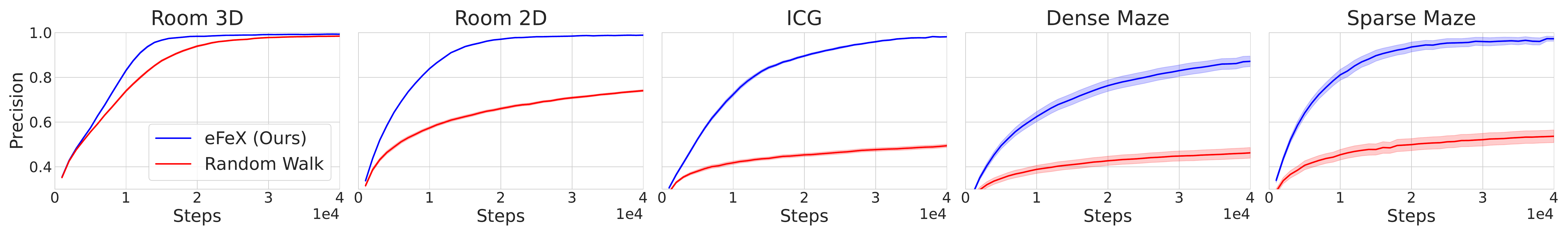}}
    \caption{Expected log-likelihood, weighted coverage, and precision, for each topology, as a function of the number of steps of the agent in the environment. See Appendix \ref{sec:app_performance} for further explanation. Higher is better. Results are averaged over 40 runs. Bands indicate 95\% confidence intervals for the averages. Random policy shown in blue, eFeX in red.}
\end{figure}

\end{document}